\pdfoutput=1
\RequirePackage{scrlfile}
\PreventPackageFromLoading{subfig}
\documentclass[11pt,a4paper]{article}
\usepackage[nohyperref]{acl2019}
\usepackage{times}
\usepackage{latexsym}
\usepackage{url}

\usepackage{graphicx}  % includegraphics, resizebox

\makeatletter
\def\adl@drawiv#1#2#3{%
        \hskip.5\tabcolsep
        \xleaders#3{#2.5\@tempdimb #1{1}#2.5\@tempdimb}%
                #2\z@ plus1fil minus1fil\relax
        \hskip.5\tabcolsep}
\newcommand{\cdashlinelr}[1]{%
  \noalign{\vskip\aboverulesep
           \global\let\@dashdrawstore\adl@draw
           \global\let\adl@draw\adl@drawiv}
  \cdashline{#1}
  \noalign{\global\let\adl@draw\@dashdrawstore
           \vskip\belowrulesep}}
\makeatother
\usepackage{booktabs} % for toprule, midrule, bottomrule
\usepackage{arydshln} % for multicolumn
\usepackage{multirow}  % for multirow
\usepackage{tabularx} % for table format such text wrapping
\usepackage{cleveref} % for refering multiple labels at once
\usepackage{subcaption}
\usepackage{longtable}
\usepackage{multicol} % for \onecolumn and \twocolumn
\usepackage[section]{placeins} % for \FloatBarrier
\usepackage{float} % for H to prevent table from floating
\usepackage{CJKutf8}
\usepackage{graphicx}
\usepackage[tableposition=below]{caption}
\usepackage{tikzsymbols}
\newcommand{\turn}[1]{``{#1}''}
\newcommand{\aturn}[1]{{\it A:} #1}
\newcommand{\bturn}[1]{{\it B:} #1}
\newcommand{\rates}[1]{{[\it #1}]}

\newcommand{\bestppl}{10.2}
\newcommand{\humanissa}{86}
\newcommand{\meenabase}{{Meena (base)}}
\newcommand{\meenafull}{{Meena (full)}}

\newcommand\blfootnote[1]{%
  \begingroup
  \renewcommand\thefootnote{}\footnote{#1}%
  \addtocounter{footnote}{-1}%
  \endgroup
}
\aclfinalcopy

\title{Towards a Human-like Open-Domain Chatbot}

\author{Daniel Adiwardana \hspace{2mm} Minh-Thang Luong \hspace{2mm} David R. So \hspace{2mm} Jamie Hall \\ \textbf{Noah Fiedel} \hspace{2mm} \textbf{Romal Thoppilan} \hspace{2mm} \textbf{Zi Yang} \hspace{2mm} \textbf{Apoorv Kulshreshtha} \\ \textbf{Gaurav Nemade} \hspace{2mm} \textbf{Yifeng Lu} \hspace{2mm} \textbf{Quoc V. Le} \\
  Google Research, Brain Team \\
  \texttt{\{adiwardana,thangluong,davidso,jamiehall,nfiedel,romzee,ziy,} \\
  \texttt{apoorvk,gnemade,yifenglu,qvl\}@google.com}
}

\begin{document}
\maketitle
\begin{abstract}
We present Meena, a multi-turn open-domain chatbot trained end-to-end on data mined and filtered from public domain social media conversations. This 2.6B parameter neural network is simply trained to minimize perplexity of the next token. 
We also propose a human evaluation metric called Sensibleness and Specificity Average (SSA), which captures key elements of a human-like multi-turn conversation. Our experiments show strong correlation between perplexity and SSA. The fact that the best perplexity end-to-end trained Meena scores high on SSA (72\% on multi-turn evaluation) suggests that a human-level SSA of \humanissa\% is potentially within reach if we can better optimize perplexity. Additionally, the full version of Meena (with a filtering mechanism and tuned decoding) scores 79\% SSA, 23\% higher in absolute SSA than the existing chatbots we evaluated.
\end{abstract}

\section{Introduction}
\begin{figure}[ht]
\centering
\includegraphics[height=150pt]{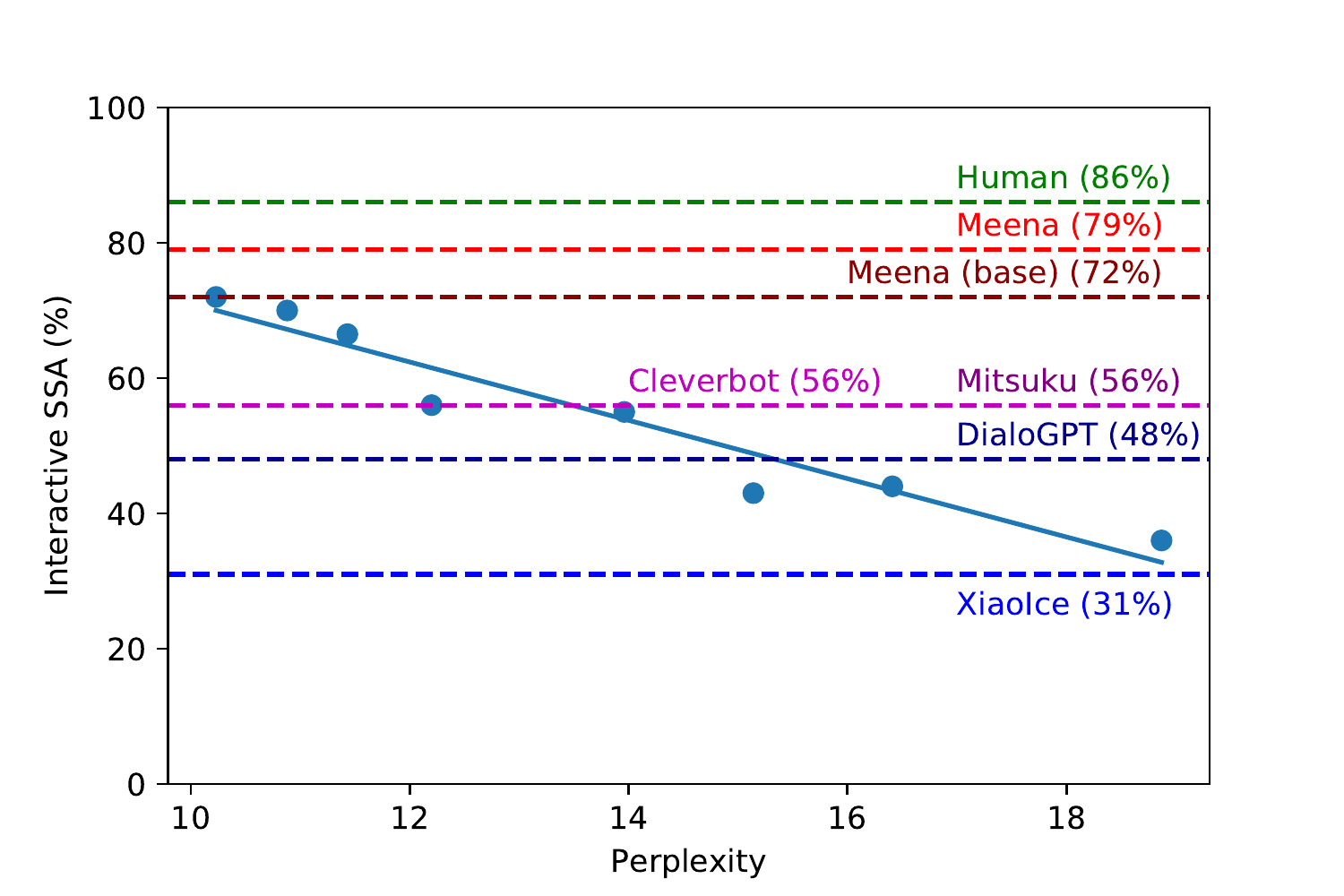}
\caption{Interactive SSA vs Perplexity. Each point is a different version of the Meena model. A regression line is plotted, for which the coefficient of determination ($R^2$) is 0.93, an indication of strong correlation between perplexity and the human evaluation metric (SSA). The dotted lines show the SSA performance of other chatbots, humans (86\%), the best end-to-end trained Meena model (72\%), and the full version of Meena which incorporates a filtering mechanism and tuned decoding (Section \ref{sec:advance_ssa}) and scores 79\%. Mitsuku and Cleverbot scored the same on overall SSA, but Mitsuku displayed higher sensibleness, whereas Cleverbot had higher specificity. See Sections \ref{sec:cleverbot_dialogpt_eval}, \ref{sec:xiaoice_mitsuku_eval}, and \ref{sec:results_other_bots} for more details on how we performed these comparisons and how to interpret the results.\label{fig:ppl-issa}}
\end{figure}

The ability to converse freely in natural language is one of the hallmarks of human intelligence, and is likely a requirement for true artificial intelligence. In order to explore this aspect of intelligence, many researchers are working on open-domain chatbots. Unlike closed-domain chatbots, which respond to keywords or intents to accomplish specific tasks, open-domain chatbots can engage in conversation on any topic. Some open-domain chatbots such as MILABOT~\cite{milabot}, XiaoIce~\cite{xiaoice}\footnote{\url{https://www.msxiaobing.com/}}, Gunrock~\cite{gunrock}, Mitsuku~\cite{mitsuku18}\footnote{\url{https://www.pandorabots.com/mitsuku/}} and Cleverbot\footnote{\url{https://www.cleverbot.com/}} (by Rollo Carpenter) display human-like attributes, but rely on complex frameworks, such as dialog managers with knowledge-based, retrieval-based, or rule-based systems. End-to-end neural network approaches \cite{shang15,vinyals15,sordoni15,serban16,dialogpt}, on the other hand, offer the simplicity of a single learned model.\blfootnote{Conversations with Meena, and with various other chatbots, are available at \url{https://github.com/google-research/google-research/tree/master/meena/}} Despite much research, open-domain chatbots still have weaknesses that prevent them from being generally useful: they often respond to open-ended input in ways that do not make sense, or with replies that are vague and generic.

Here we present Meena, a generative chatbot model that was trained end-to-end on 40B words mined and filtered from public domain social media conversations. With Meena, we push the limits of the end-to-end approach and show that a large-scale low-perplexity model can be a good conversationalist. We use a seq2seq model~\citep{sutskever14,bog15} with the Evolved Transformer~\citep{et} as the main architecture. The model is trained on multi-turn conversations where the input sequence is all turns of the context (up to 7) and the output sequence is the response. Our best model has 2.6B parameters and achieves a test perplexity of \bestppl{} based on a vocabulary of 8K BPE subwords~\cite{senrich15}.

To measure the quality of Meena and other chatbots, we propose a simple human evaluation metric. Sensibleness and Specificity Average (SSA) combines two fundamental aspects of a human-like chatbot: making sense and being specific. We ask human judges to label every model response on these two criteria. The first part of the metric, sensibleness, is a basic requirement. To converse properly with a human, a bot's responses have to make sense in context; humans typically take this for granted when conversing with one another, and our evaluations find that 97\% of  human-produced statements meet this criterion (see Section~\ref{sec:human_level}). However, making sense is not enough. If a model is designed with sensibleness as its only objective, its responses could be vague and boring, since that is a safe strategy to avoid being penalised for not making sense. For example, closed-domain chatbots typically respond with a generic apology when a human asks something outside their domain; some end-to-end learned chatbots respond ``I don't know'' to many inputs \citep{li16diversity}; and Turing Test contest entrants often try to avoid detection by being strategically vague \citep{amazoneval}. They succeed in not generating gibberish or contradicting themselves, but at the cost of not really saying anything of substance. To mitigate this, we add a second dimension to the SSA metric, which asks our evaluators whether a response is specific given the context. This prevents bots from hiding behind vague replies, allowing us to more openly examine what they are capable of. As discussed in Section~\ref{subsec:metrics}, this successfully distinguishes between generic and lively responses, while also being simple and easy for crowd workers to understand.

We compare Meena, humans, and other open-domain chatbots using the SSA metric with two types of human evaluation: \textit{static} and \textit{interactive}. For static evaluation, we curated a dataset with 1,477 multi-turn conversations. For interactive evaluation, humans could chat about anything they wanted. We were surprised, but pleased, to discover that the SSA metric shows strong correlation with Meena's perplexity, both in static and interactive evaluation. In other words, the better that Meena fit its training data, the more sensible and specific its chat responses became. At first glance, this result may seem intuitive, but it surprised us because recent research found a poor correlation between human evaluation scores and automatic metrics such as BLEU \citep{liu16,lowe17}.

Our best end-to-end learned model has an average of 72\% SSA. The full version of Meena scores 79\% by incorporating a filtering mechanism and tuned decoding (Section \ref{sec:advance_ssa}). This is still below the \humanissa\% SSA achieved by an average human, but is far closer than the other chatbots we tested. We note that humans have very high sensibleness, but significantly lower specificity, as detailed in Section \ref{sec:human_level}.

We will also discuss weaknesses of our methodology. For example, our static evaluation dataset is too restricted to capture all aspects of human conversations. Nevertheless, the fact that Meena achieves such a high SSA score and that there is a correlation between SSA and perplexity means that a human-like chatbot, in terms of sensibleness and specificity, could be in sight if we can attain better perplexity.

Our contributions are: (1) proposing a simple human evaluation metric for multi-turn open-domain chatbots that captures basic, but important, attributes of human conversation; (2) showing evidence that perplexity is an automatic metric that correlates with human judgment, in contrast to recent findings on other automatic metrics mentioned above; (3) demonstrating that an end-to-end neural model with sufficiently low perplexity can surpass the sensibleness and specificity of existing chatbots that rely on complex, handcrafted frameworks developed over many years.

\section{Evaluating chatbots}
\label{sec:evaluation}
Evaluating chatbots and natural language generation is a well-known challenge \citep{liu16, lowe17, novikova17, hashimoto19}, which we aim to address in this paper. First, we propose a human evaluation metric that captures key elements of human-likeness of conversational responses (Section 2.1). We then describe two human-evaluation setups: {\it static}, in which we benchmark models on a fixed set of multi-turn contexts to generate responses (Section 2.2); and {\it interactive}, where we allow humans to chat freely with chatbots (Section 2.4). Lastly, we detail our automatic evaluation metric for fast development and end-to-end optimization (Section 2.7).

\subsection{Measuring Human Likeness}
\label{subsec:metrics}

To measure the quality of a response given a context, we propose a sequence of two questions. We first ask whether the response, given the context, makes sense. Sensibleness arguably covers some of the most basic aspects of conversational human-likeness, such as common sense and logical coherence. Sensibleness also captures other important aspects of a chatbot, such as {\it consistency}. The crowd worker is asked to use common sense to judge if a response is completely reasonable in context. If anything seems off — confusing, illogical, out of context, or factually wrong — then it should be labeled as, ``does not make sense''.

However, being sensible is not enough. A generic response (e.g., \textit{I don't know}) can be sensible, but it is also boring and unspecific. Such responses are frequently generated by bots that are evaluated according to metrics like sensibleness alone \citep{li16diversity,amazoneval}. To illustrate this, we create GenericBot: a trivial bot that always replies to questions with ``I don't know" and to statements with ``ok" (examples in Appendix Table \ref{t:boring_bot_static_analysis}). On \textit{static} evaluation (using a fixed set of prompts and bot-generated responses), 70\% of GenericBot's responses are labeled sensible, surpassing even DialoGPT (62\%), even though DialoGPT is clearly more human-like than GenericBot. To overcome this issue, we need our evaluation to separate more fully human-like conversation from bland and generic statements. 
Therefore, if a response is labeled as sensible, we further ask the crowd worker to determine if it is specific to the given context. For example, if A says, ``I love tennis,'' and B responds, ``That's nice,'' then the utterance should be marked, ``not specific''. That reply could be used in dozens of different contexts. However, if B responds, ``Me too, I can't get enough of Roger Federer!'' then it is marked as ``specific'', since it relates closely to what is being discussed. Responses labeled not sensible are considered not specific. In GenericBot's case, none of the responses are specific, whereas 39\% of DialoGPT's responses are specific.

This sequence of two questions is designed to start with the most concrete and basic human quality (sensibleness) and then progress to the arguably more subjective human quality (specificity). The degree of subjectivity is somewhat quantified in the crowd worker agreement. We measure crowd worker consistency for every model benchmark using agreement and Krippendorff's alpha~\cite{krippendorff15}, shown in Table \ref{t:rater_consistency}. The agreement is reasonable considering the questions are subjective and the final results are always aggregated labels (e.g., average sensibleness across all chatbot responses).

\begin{table}[h!]
\small
\centering
\begin{tabular}{c|c|c}
\toprule
{Metric} & {\bf Agreement (\%)} & {\bf Krippendorff's alpha} \\
\midrule
Sensibleness & $76 \pm 3$ & $0.42 \pm 0.03$ \\
\midrule
Specificity & $66 \pm 2$ & $0.30 \pm 0.05$  \\
\bottomrule
\end{tabular}
\caption{The average and standard deviation of crowd worker agreement across static evaluations of Meena models. Each static evaluation consisted of 1,477 $(context, response)$ pairs, each labeled by 5 crowd workers.}
\label{t:rater_consistency}
\end{table}

Given a set of responses labeled as described above, we can calculate sensibleness and specificity as the percentage of responses labeled as sensible and specific, respectively. To combine these two into one metric, we take a simple average of the two, which we call SSA (\textbf{s}ensibleness and \textbf{s}pecificity \textbf{a}verage). SSA is a proxy for human likeness, which also penalizes chatbots that consistently produce generic responses. For example, GenericBot's SSA is 35\% and DialoGPT's SSA is 51\%, providing a much more fair separation and ranking than sensibleness alone.

Before arriving at SSA, and before any of the chatbots were tested, the authors of this paper conducted several rounds of pilot studies on what to ask crowd workers and how to best phrase the instructions. We settled on the two-question SSA for several reasons: it was easy for crowd workers to understand; alternative additional questions did not add extra information; and more subjective questions result in lower agreement between crowd workers.

\begin{figure}
\centering
\includegraphics[height=150pt]{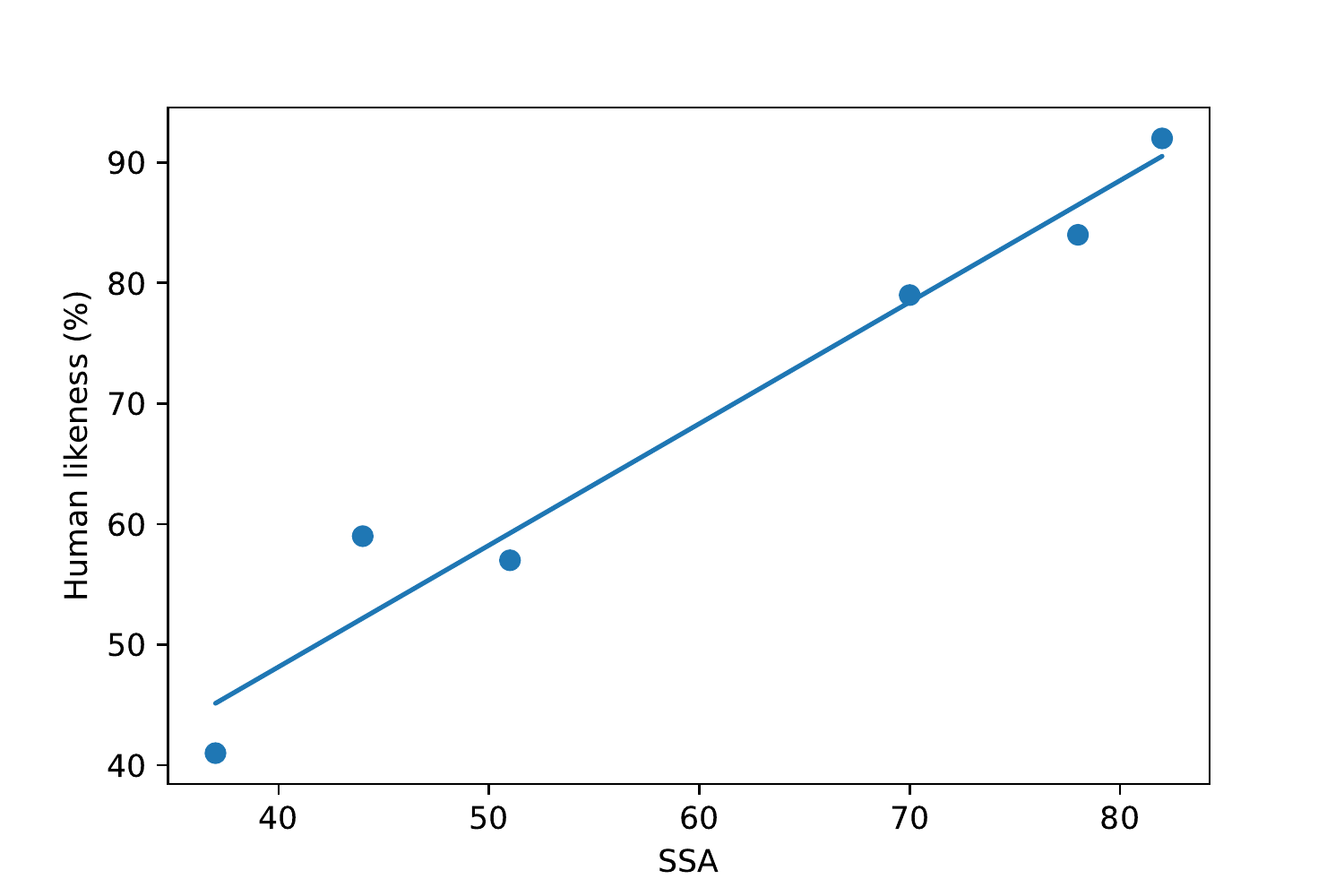}
\caption{SSA vs human likeness. Each point is a different chatbot, except for the top right one, which is human. A regression line is plotted, for which the coefficient of determination ($R^2$) is 0.96. The SSA values were collected using static evaluation mode (Section \ref{subsec:ssa}). The human likeness evaluation was also conducted in static evaluation mode. Instead of judging sensibleness or specificity, however, we asked crowd workers to judge whether a given response was ``human-like", or in other words, looked like a response that a human might give in the provided context.
}
\label{fig:ssa-human-likeness}
\end{figure}

As an additional check on the SSA metric, we reran a static evaluation, this time asking crowd workers to assess whether or not a response is ``humanlike''. We find that there is a high correlation between those labels and the two components of the SSA metric (Figures \ref{fig:ssa-human-likeness}, \ref{fig:sensibleness-human-likeness}, \ref{fig:specificity-human-likeness}). Compared to a direct evaluation of what crowd workers consider to be ``humanlike'', SSA has significant advantages for large-scale evaluation tasks: it is more objective, easier for crowd workers to understand, and penalizes boring and vague responses. Nevertheless, these findings give us confidence that SSA is indeed capturing important aspects of human likeness.

\subsection{Static Evaluation}
\label{subsec:ssa}
In order to have a common benchmark to easily compare models, we create a collection of
1,477 conversational contexts with between 1 and 3 conversation turns, that we call the Mini-Turing Benchmark (MTB).
We started this dataset by compiling single-turn contexts (e.g., ``How are you?") from multiple sources, such as from the work\footnote{\url{http://ai.stanford.edu/~quocle/QAresults.pdf}} of~\citet{vinyals15} and the transcripts of the Loebner Prize\footnote{\url{https://aisb.org.uk/events/loebner-prize}} contests (years 2014-2018).
In total, there were 315 single-turn contexts, which we then extended to include 500 two-turn and 662 three-turn contexts.
 
The MTB also contains contexts with personality questions (e.g. ``Do you like cats?''), some of which expect responses with personality consistency. For example, the context ``A: Do you like movies?; B: Yeah. I like sci-fi mostly; A: Really? Which is your favorite?'' expects a consistent response such as \textit{I love Back to the Future}. On the other hand, a response like \textit{I don't like movies} would be a contradiction, and thus not considered sensible.

When evaluating chatbots, all MTB contexts are fed to the models or presented to humans to obtain responses. We send the resulting $(context, response)$ pairs to crowd workers and asked whether each response given the context is sensible and specific as defined in \ref{subsec:metrics}. We call this {\it static evaluation} because the contexts are fixed.

\subsection{Interactive Evaluation}
\label{subsec:interactive}
Static evaluation may be suitable for comparing models, but it is biased by how the static evaluation dataset was constructed. To address this, we create an additional evaluation mode where the crowd workers can chat 1:1 with a chatbot about anything they want. As with static evaluation, workers are also asked to decide whether each response from the chatbot is sensible and specific as defined in \ref{subsec:metrics}. Conversations start with ``Hi!'' from the chatbot to mark the beginning of the conversation and crowd workers have no expectation or instructions about domain or topic of the conversation. A conversation is required to last at least 14 turns (7 from chatbot) and at most 28 turns. We collected 100 such conversations for each model (i.e., at least 700 labeled turns per model). We then measure the percentage of labeled turns that are sensible and specific.

Unlike a typical Turing test~\citep{turing50}, we tell the human judges upfront that they are about to chat with an experimental chatbot and ask them to label what the chatbot says in terms of sensibleness and specificity. This shifts the focus of the judges and chatbot creators from optimizing for deception detection to optimizing for detecting and maximizing human-like qualities (e.g., sensibleness). Similar to our approach, \citet{asma19} also conduct interactive evaluation by allowing humans to chat freely with bots. Their setup, however, focuses on evaluating conversations as a whole (as opposed to at the level of individual turns) and judges evaluate for quality, fluency, diversity, relatedness, and empathy.

\subsection{Estimate of Human Performance}
To estimate static SSA of humans we ask crowd workers to respond to MTB contexts. Additionally, to estimate human interactive SSA, we leveraged the help of internal company volunteers to collect 100 human-human conversations following mostly the same instructions as crowd workers for every other chatbot. Labeling of sensibleness and specificity was conducted by independent crowd workers with majority voting of 5 workers per human turn. The difference from the rest of the evaluations is that, in this case, participants knew they were chatting with another human. In contrast, when humans chat with a chatbot they will occasionally say unusual things to test the chatbot's limits. \citet{hill15} describe differences in human behavior when talking to a chatbot. That said, we never incentivize humans to chat adversarially with chatbots in any of our evaluations.

\subsection{Evaluation of Cleverbot and DialoGPT}
\label{sec:cleverbot_dialogpt_eval}
To integrate with Cleverbot, we leverage its API. For DialoGPT, we use its open sourced 762M parameter model.\footnote{\url{https://github.com/microsoft/DialoGPT}} It is worth mentioning that we initially tried the 345M parameter DialoGPT model, because it was reported to perform best on single-turn human evaluation. However, the 345M parameter model seemed to perform noticeably worse than the 762M one in preliminary evaluations of multi-turn conversations. Our human evaluation is multi-turn, so we select the 762M model.

The DialoGPT authors were unable to release their decoding script at the time of writing. Therefore, following their published description, we use top-K decoding with $K=10$. We adapt the decoding implementation by \citet{wolf19}. Moreover, since the backward model was also not released we were not able to try their MMI re-ranking~\citep{li16diversity}.

Both Cleverbot and DialoGPT were evaluated using the same crowd sourcing setup as for Meena.

\subsection{Evaluation of Mitsuku and XiaoIce}
\label{sec:xiaoice_mitsuku_eval}
Because we chose to use the free Mitsuku web app\footnote{Pandorabots offers a paid enterprise package, which includes the Mitsuku API.}, and there is no public API for XiaoIce, we called on the help of internal company volunteers and only conducted interactive evaluation. Volunteers collectively had 100 conversations with Mitsuku, and 119 with XiaoIce on their publicly available web apps. The volunteers conversed with the chatbots following mostly the same instructions that crowd workers follow for every other chatbot. The difference is that humans would say ``Hi!" for the first turn, instead of the chatbot, in order to keep the first turn the same as other cases. Labeling of sensibleness and specificity in all cases was conducted by independent crowd workers with majority voting of 5 workers per chatbot turn. Also note that both XiaoIce and Mitsuku sometimes include an image in their reply and occasionally, volunteers include text descriptions of the images they see. The presence of the image may in some cases change the sensibleness of the response for better or worse.

XiaoIce interacts in Mandarin so both the volunteers and the independent crowd workers were native Mandarin speakers. The group of volunteers for XiaoIce, Mitsuku, and human-human conversations were mostly disjoint. Other than requiring a knowledge of Mandarin for XiaoIce conversations, volunteer selection was arbitrary. We had 29 volunteers for XiaoIce, 43 for Mitsuku, and 21 for human-human.

To reset Mitsuku state between conversations, volunteers refreshed the web page. During the writing of this paper there was no clear way to reset the state of XiaoIce. The XiaoIce team have informed us that not resetting the state negatively affects the model's control of the context.\footnote{From personal communication with the XiaoIce team, after the writing of the paper.} Also, most XiaoIce volunteers shared the same Weibo account.\footnote{Weibo is a microblogging service mostly used in China, which also allows users to chat with XiaoIce: \url{https://www.weibo.com/}} The XiaoIce team confirmed that account reuse negatively impacts the internal profile constructed by XiaoIce for a user. The XiaoIce team further suggested that, if the same Weibo account needs to be reused, we should wait at least one hour between volunteers using the account. In our experiments, we may have sometimes waited less than that amount of time between volunteers, although we made sure the account was only used by one volunteer at a time. Finally, the XiaoIce team mentioned that in the past few months (as of this writing), a limited version of XiaoIce with the smallest index has been served on Weibo. This version is expected to produce less satisfactory responses.

Direct comparisons between XiaoIce and other chatbots come with a caveat: XiaoIce can be seen as a product that optimizes for long-term user engagement, of which dialog generation is just one component. In other words, Meena is arguably at an advantage when comparing SSA scores.

\subsection{Automatic Evaluation}
\label{subsec:auto}
For quick research iterations, we focus on perplexity. Unlike the previous two evaluation types, perplexity is an automatic metric. A seq2seq model outputs a probability distribution over possible next response tokens. Perplexity measures how well the model predicts the test set data; in other words, how accurately it anticipates what people will say next. When interpreting perplexity scores, bear in mind that lower is better and that the theoretical minimum is one.

As shown in Section \ref{sec:results}, this commonly used metric correlates with human judgement of sensibleness and specificity. This is encouraging, because it is both automatic and directly optimizable with the standard cross-entropy loss function.

\section{Meena chatbot}
\label{sec:meena}
\begin{table*}[tbh!]
\small
\centering
\begin{tabular}{p{12cm}|>{\centering\arraybackslash}p{1cm}}
\toprule
\multicolumn{1}{c|}{{\bf Responses} for \turn{Why do you like the ocean?}} & {\bf Score} \\
\midrule
1. I love the beach. I like that it doesn't have anything in it. Just open water. & -12.98 \\
2. It's dark and mysterious. I discovered it when I was really young and it was a dream of mine since. The ocean is full of wonders, mysteries and great things to discover. I love snorkeling. I moved to Hawaii a year ago and I'm going to college for marine biology. So there's that. & -14.03 \\
3. The sea sounds cool and it's really satisfying to swim in. & -14.62 \\
4. I dunno. It's beautiful, I guess. It reflects my upbringing, I reckon. & -14.87 \\
5. Well I'm from the pacific northwest and always somehow end up back there. I've been to Hawaii but I'm not one for the beach, it's too hot and the air is so dry. & -15.36 \\
6. The beauty of it is incredible, and swimming in it and being able to see huge waves sweeping by is just amazing. I was born in the desert so it's really the only place in the world I've seen with flowing water around me & -17.07 \\
7. It's beautiful, maybe? I just wish I lived closer to it! & -17.14 \\
8. It's pretty and soothing. & -17.19 \\
9. Because I love the smell. I love walking when I can through the beach. I live in California so I basically go to the ocean almost every weekend. & -17.38 \\
10. The ocean is pretty cool, I suppose. I like the various types of fish. & -17.90 \\
\bottomrule
\end{tabular}
\caption{{\bf Sampling Outputs} -- responses found by sampling with temperature, sorted by the length-normalized log-likelihood scores and using the same Evolved Transformer as the one used in Table~\ref{t:beam}. The scores are calculated as $\frac{\log P}{T}$, where $P$ is the likelihood of the response and $T$ is the number of tokens.}
\label{t:sampling}
\end{table*}

\begin{table}
\small
\centering
\begin{tabular}{p{6cm}|c}
\toprule
\multicolumn{1}{c|}{{\bf Responses} for \turn{Why do you like the ocean?}} & {\bf Score} \\
\midrule
1. Why don't you like the ocean? & -1.70 \\
2. I don't like the ocean. I love the ocean. & -2.66 \\
3. I don't like the ocean. I love it. & -2.78 \\
4. I don't like the ocean, I love the ocean. & -2.94 \\
5. I don't like the ocean, I love it. & -2.94 \\
6. Why don't you like the ocean? :P & -2.95 \\
7. I don't like the ocean, I love it! & -3.15 \\
8. I don't like the ocean. I love the ocean! & -3.20 \\
9. Why don't you like the ocean? It's beautiful. & -3.26 \\
10. I don't like the ocean. I love the ocean. There's a difference. & -3.31 \\
\bottomrule
\end{tabular}
\caption{{\bf Beam Search Outputs} -- top responses generated by beam-search decoding and the corresponding length-normalized log-likelihood scores. We use an Evolved Transformer with perplexity 10.2 and vocabulary size of 8K.}
\label{t:beam}
\end{table}

As described above, recent work on end-to-end dialog models has fallen into two broad categories: (1) complex models with human-designed components, and (2) large neural network models (known as end-to-end models) that are closer to generic learning frameworks. End-to-end models have shown promise, but clear limitations \cite{gao19}. An open question has been: in order to reach a point where a model can carry out high-quality, multi-turn conversations with humans, could we simply take an end-to-end model and make it bigger---by adding more training data and increasing its parameter count---or is it necessary to combine such a model with other components? In this section we describe the Meena model, the largest end-to-end model to enter the field so far. We believe it answers the open research question, by showing that a large end-to-end model can generate almost humanlike chat responses in an open-domain setting.

In this section, we will describe the training data, architecture, and decoding algorithm. We will also provide a few sample conversations that Meena has had with humans. 

\subsection{Training Data}
\label{sec:dataset}
The dataset used to train Meena is mined and filtered from public domain social media conversations.
The source data are essentially message trees involving multiple speakers: the very first message is the root; replies to a message are its child nodes. Any path along the tree induces a conversation where each message is a conversation turn. By treating each turn in a conversation path as a response and all the previous turns (up to 7) as a context, we create a training example of the form {\it (context, response)} pair.

We also filter the data to improve the generation quality.
A message is removed if any of the following conditions holds: 1.~the number of subwords is less than 2 or more than 128; 2.~the percentage of alphabetic characters is less than 70\%; 3.~message contains URL; 4.~author's username contains ``bot''; 
5.~the message is repeated more than 100 times;
6.~the message has a high $n$-gram overlap with the parent's text;
7.~the message is potentially unsafe or offensive with respect to a commercial text classifier.
In addition, we remove copies of the parent's text quoted in a message.

For simplicity, when a message is removed, we drop all sub-trees rooted under it. 
After these filtering steps, the number of $(context, response)$ pairs extracted is 867M. The text is tokenized using byte-pair-encoding (BPE)~\cite{senrich15} with the sentencepiece library.\footnote{\url{https://github.com/google/sentencepiece}} We use a vocabulary of 8K BPE subwords, which we found in our early experiments to be sufficient for generating specific responses while still allowing us to fit larger models in memory.

The final Meena dataset contains 341GB of text (40B words). In comparison, GPT-2 \cite{gpt2} has been trained on 40GB of Internet text (8 million web pages).

\subsection{Model Architecture}
The best performing Meena model is an Evolved Transformer (ET)~\citep{et} seq2seq model with 2.6B parameters, which includes 1 ET encoder block and 13 ET decoder blocks. The Evolved Transformer is an evolutionary NAS architecture~\citep{real17,real18} based on the Transformer~\cite{transformer}. Our largest (i.e., maximum memory usage) Evolved Transformer scored 10.2 perplexity and our largest vanilla Transformer scored perplexity 10.7 for the same number of training steps (738k). The largest vanilla Transformer had 32 decoder layers with other architectural hyperparameters held constant.\footnote{An Evolved Transformer block is about twice as deep as a Transformer layer}

For comparison, the extra-large GPT-2 model~\cite{gpt2} has 1.5B parameters and is a language model (i.e., decoder only); whereas the large conversational model from the recent DialoGPT work~\cite{dialogpt} has 762M parameters.

Meena's hidden size is 2,560 and the number of attention heads is 32. We share the embeddings across the encoder, the decoder, and the softmax layer. The encoder and decoder each have a maximum length of 128 tokens (i.e., 256 combined). The hyperparameters of our best model were found via manual coordinate-descent search.

\subsection{Training Details}
We trained our best model for 30 days on a TPU-v3 Pod (2,048 TPU cores) on the Meena dataset containing 40B words (or 61B BPE tokens). Interestingly, the 2.6B-parameter model can overfit~\footnote{In the sense that validation loss increases as train loss decreases.} on a 61B-token dataset which suggests a surprisingly large model capacity.
Therefore, we add a small amount of 0.1 attention and feed-forward layer dropout. Additionally, to save memory, we chose the Adafactor optimizer~\cite{shazeer18} with 0.01 as the initial learning rate, keeping it constant for the first 10k steps and then decaying with the inverse square root of the number of steps. We use the Tensor2Tensor codebase~\cite{tensor2tensor} for training Meena.\footnote{\url{https://github.com/tensorflow/tensor2tensor}}

A TPU-v3 core has 16GB of high-bandwidth memory. We maximized memory usage for model parameters and stored only 8 training examples per core. Each training step took about 1 second. In the full TPU-v3 Pod, this meant we learned over 4M tokens per training second. Therefore, by the end of training, the model had traversed the full training set 164 times (or epochs) and observed a total of about 10T tokens (including repeated ones).

\subsection{Decoding}
\label{meena-decoding}

Generating generic (i.e., not specific) and bland responses \citep{li16diversity} has always been a major challenge in existing neural conversational models. A common approach to mitigating this problem is to use more sophisticated decoding algorithms, for instance with different forms of re-ranking \citep{li16diversity,shao17} or conditioning on profiles, topics, and styles \citep{li16,wang17,xing17,zhang18personal}. Recent works also explore new frameworks such as adversarial learning \citep{li17adversarial,zhang18}, variational autoencoding \citep{zhao17,gu19}, or both \citep{gao19joint} at the cost of added complexity and less scalability.

In contrast, we show that given a model with sufficiently low perplexity, a simple sample-and-rank decoding strategy achieves both diverse and high-quality responses. Sample-and-rank, works as follows: First, we sample $N$ independent candidate responses using plain random sampling with temperature $T$. Second, we select the candidate response with the highest probability to use as the final output.

Temperature $T > 0$ is a hyper-parameter that regulates the probability distribution $p_i$ of the next token during decoding. We divide the logits $z_i$ by $T$ before computing the ``softmax" as in \citet{hinton15}:

\begin{equation}
p_i = \frac{\exp(z_i/T)}{\sum_{j}\exp(z_j/T)}
\end{equation}

$T=1$ yields the unmodified distribution. We observe that large values of $T$ favor contextually rare tokens, such as relevant entity names, but might also assign too much probability to incorrect tokens depending on the model's predictions. Meanwhile, smaller values of $T$ favor more common words such as articles or prepositions, which are safer but less specific.

Tables~\ref{t:sampling} and \ref{t:beam} show responses for the arbitrary probing input \turn{Why do you like the ocean?} under sample-and-rank and beam-search, respectively.
As we can see, beam-search decoding generates repetitive and uninteresting responses. On the other hand, sample-and-rank provides us with diverse and content-rich responses. The key here is to have a model with low perplexity so samples can be taken at high temperature to produce human-like content.

For all the results in Section~\ref{sec:results}, we use sample-and-rank with $N=20$ and $T=0.88$. Additionally, as shown in Figure~\ref{fig:ppl-issa}, for this fixed decoding strategy, sensibleness and specificity improve as model test set perplexity falls. For additional decoding results and samples, see Section~\ref{sec:advance_ssa}.

\FloatBarrier
\subsection{Sample conversations}
\label{sec:sample_conversations}
Below are cherry picked conversations that Meena has had with humans. We selected these conversations \textit{after} they were completed. That is, the Meena responses within the conversations were not cherry picked; they were produced automatically using sample-and-rank (Section \ref{meena-decoding}). Conversations B and C are excerpts from conversations number 43 and 48, respectively, of the conversations dataset published on GitHub.\footnote{\url{https://github.com/google-research/google-research/tree/master/meena/}}

\vspace{3em}
\begin{table}[ht!]
\small
\centering
\begin{tabular}{rp{6cm}}
\toprule
\multicolumn{2}{c}{{\bf Conversation A}} \\
\midrule
\textbf{Human}: & \textit{say you're in an empty giant office and you see a black bird flying indoors. what do you do?} \\
\textbf{Meena}: & \textit{What kind of bird? Is it friendly? How large? How big is the office? Is it indoors or outdoors?} \\
\textbf{Human}: & \textit{well it's a predator bird and it's gonna eat you} \\
\textbf{Meena}: & \textit{I would grab my phone and call 911.} \\
\bottomrule
\end{tabular}
\caption*{Meena handles arbitrary open-domain input. However, asking ``Is it indoors or outdoors" was not sensible. \label{t:conv4}}
\end{table}

\begin{table}[ht]
\small
\centering
\begin{tabular}{rp{6cm}}
\toprule
\multicolumn{2}{c}{{\bf Conversation B}} \\
\midrule
\textbf{Human}: & \textit{Hi!} \\
\textbf{Meena}: & \textit{Hi! How are you this fine evening?} \\
\textbf{Human}: & \textit{I'm doing well. it's morning my time!} \\
\textbf{Meena}: & \textit{Good morning! What are your plans for today?} \\
\textbf{Human}: & \textit{Get my work done then watch a movie at home in the night} \\
\textbf{Meena}: & \textit{Sounds fun! What movie?} \\
\textbf{Human}: & \textit{I haven't decided yet, perhaps an Oscar-nominated one. any suggestion?} \\
\textbf{Meena}: & \textit{I'm not too familiar with recent Oscar winners, but I do know I wanted to see The Grand Budapest Hotel.} \\
\textbf{Human}: & \textit{ah, tell me about that movie!} \\
\textbf{Meena}: & \textit{It's from the same director and writer as Moonrise Kingdom. Wes Anderson is pretty well known for making funny, colorful films. I've heard it's pretty good.} \\
\textbf{Human}: & \textit{ha! I've searched on the internet and both movies are very colorful!} \\
\textbf{Meena}: & \textit{I'm looking forward to it. I love colorful movies.} \\
\bottomrule
\end{tabular}
\caption*{The Grand Budapest Hotel and Moonrise Kingdom are indeed colorful movies, directed by Wes Anderson.\label{t:conv1}}
\end{table}

\begin{table}[ht!]
\small
\centering
\begin{tabular}{rp{6cm}}
\toprule
\multicolumn{2}{c}{{\bf Conversation C}} \\
\midrule
\textbf{Human}: & \textit{Haha, I think mathematics has a presence beyond human existence.} \\
\textbf{Meena}: & \textit{So does philosophy.} \\
\textbf{Human}: & \textit{Can you state with examples how this is true.} \\
\textbf{Meena}: & \textit{Can you state how it isn't?} \\
\textbf{Human}: & \textit{Without human existence there will be no one left to comprehend whats right whats wrong, and thus philosophy wi [sic]} \\
\textbf{Meena}: & \textit{That's not the definition of philosophy.} \\
\textbf{Human}: & \textit{Without human existence there will be no one left to comprehend whats right whats wrong, and thus philosophy will not be needed. How do you define philosophy?} \\
\textbf{Meena}: & \textit{The study of general knowledge or the fundamental nature of things.} \\
\bottomrule
\end{tabular}
\caption*{Meena tries to define philosophy.\label{t:conv3}}
\end{table}

\begin{table}[ht!]
\small
\centering
\begin{tabular}{rp{6cm}}
\toprule
\multicolumn{2}{c}{{\bf Conversation D}} \\
\midrule
\textbf{Human}: & \textit{I like how cute cows are} \\
\textbf{Meena}: & \textit{Ah, interesting. You like how they are cute, right?} \\
\textbf{Human}: & \textit{yeah, and they're realy smart} \\
\textbf{Meena}: & \textit{I heard they go to college} \\
\textbf{Human}: & \textit{cows go to college?} \\
\textbf{Meena}: & \textit{I heard that a cow went to Harvard.} \\
\textbf{Human}: & \textit{what did the cow study?} \\
\textbf{Meena}: & \textit{Bovine sciences.} \\
\textbf{Human}: & \textit{do horses go to Harvard?} \\
\textbf{Meena}: & \textit{Horses go to Hayvard.} \\
\textbf{Human}: & \textit{that's a pretty good joke, I feel like you led me into it} \\
\textbf{Meena}: & \textit{You were trying to steer it elsewhere, I can see it.} \\
\bottomrule
\end{tabular}
\caption*{Meena executes a multi-turn joke in an open-domain setting. We were unable to find this in the data.\label{t:conv2}}
\end{table}

\FloatBarrier

Appendix \ref{apdx:conversations} shows random samples of conversations.

\section{Results}
\label{sec:results}
In this section, we will first demonstrate the correlation between test perplexity and the human evaluation metric, SSA, defined earlier. We also include human-level upperbound estimates for both static and interactive evaluations, beside performances of other chatbots, such as XiaoIce, Mitsuku, DialoGPT, and Cleverbot. Lastly, we provide sample responses for different models given the same contexts to understand how Meena qualitatively compares to others.

\begin{figure}[tbh!]
\centering
\begin{minipage}{0.45\textwidth}
  \centering
  \includegraphics[height=150pt]{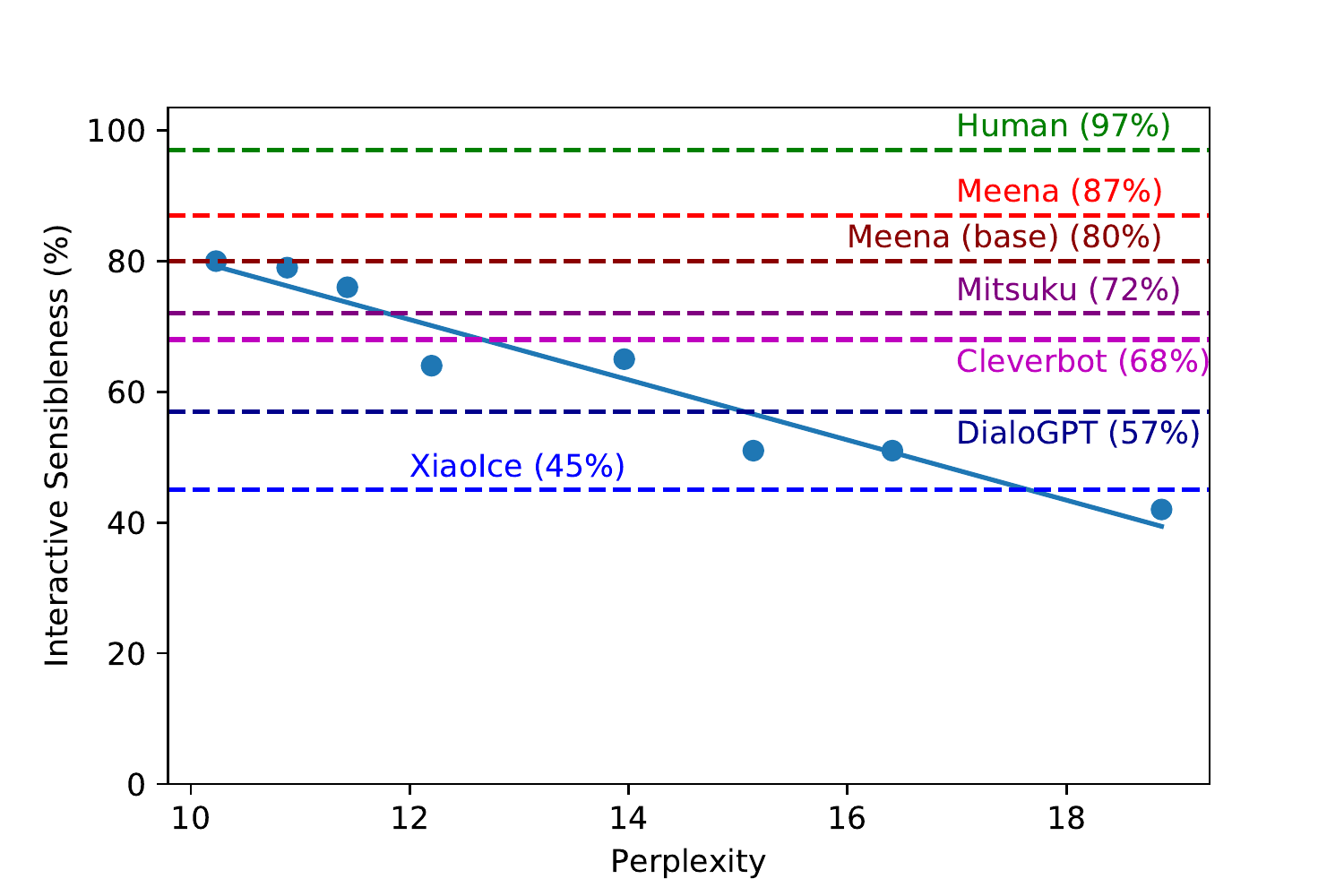}
  \caption{Interactive sensibleness vs perplexity.\label{fig:ppl-isensibleness}}
\end{minipage} \hfill
\begin{minipage}{0.45\textwidth}
  \centering
  \includegraphics[height=150pt]{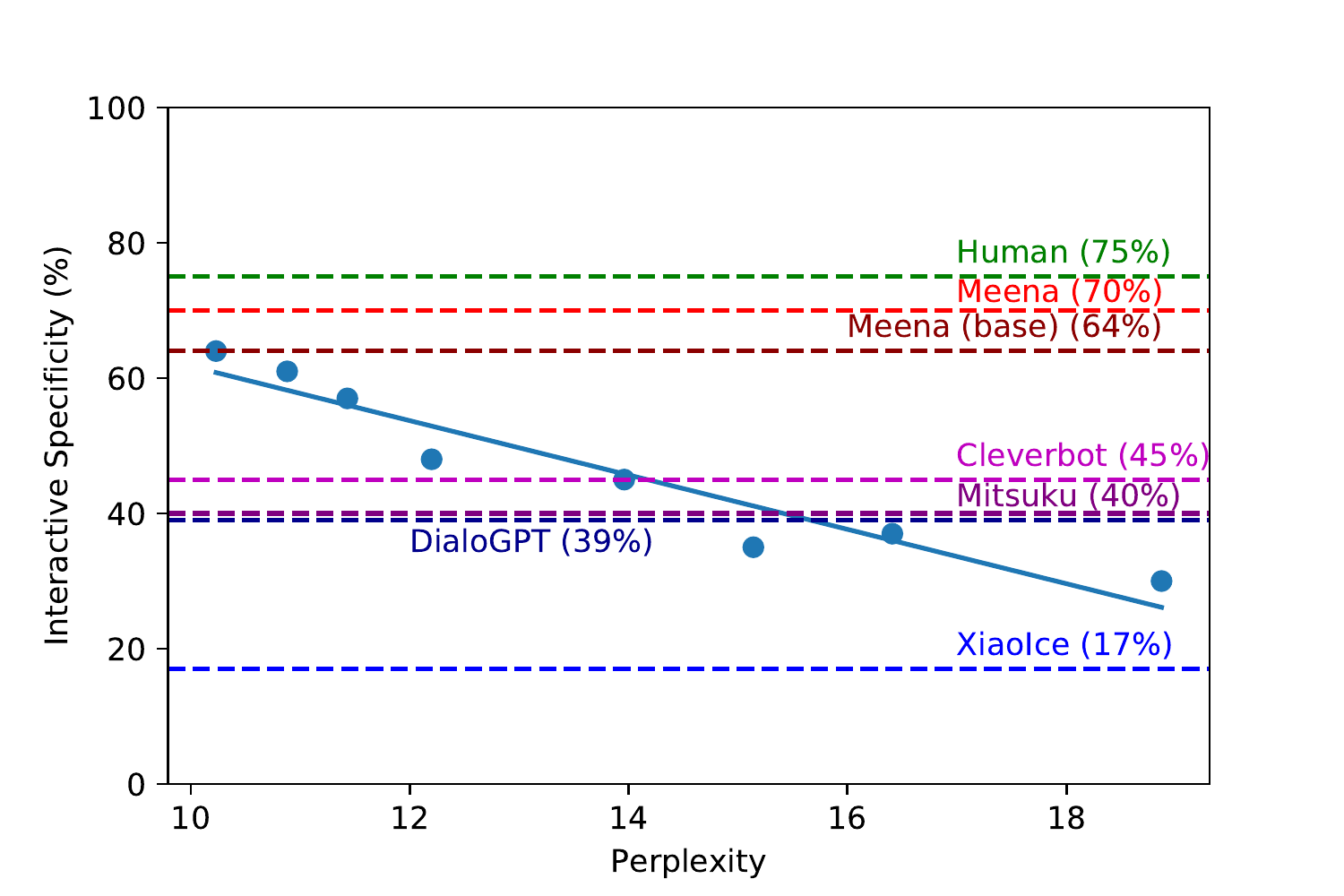}
  \caption{Interactive specificity vs perplexity.\label{fig:ppl-ispecificity}}
\end{minipage}
\end{figure}

\subsection{SSA-perplexity correlation}
We trained models with different hyper-parameter settings and architectures on the dataset described in Section \ref{sec:dataset}. We vary the number of layers, attention heads, total training steps, whether we use Evolved Transformer or regular Transformer and whether we train with hard labels or soft labels/distillation \cite{hinton15}. The trained models are then measured with an automatic metric, test perplexity (Section \ref{subsec:auto}), and also with human metrics (\Cref{subsec:ssa,subsec:interactive}). Our results indicate most of the variance in the human metrics can be explained by the test perplexity. The end-to-end trained Meena model with lowest perplexity is referred to as {\it \meenabase{}}. In addition, we also include an improved version of Meena (detailed in Section 5) and refer to this as the {\it{\meenafull{}}} model, or just {\it{Meena}} model for short.

\begin{figure*}[tbh!]
    \begin{subfigure}[t]{0.32\textwidth}
        \includegraphics[width=1.0\textwidth]{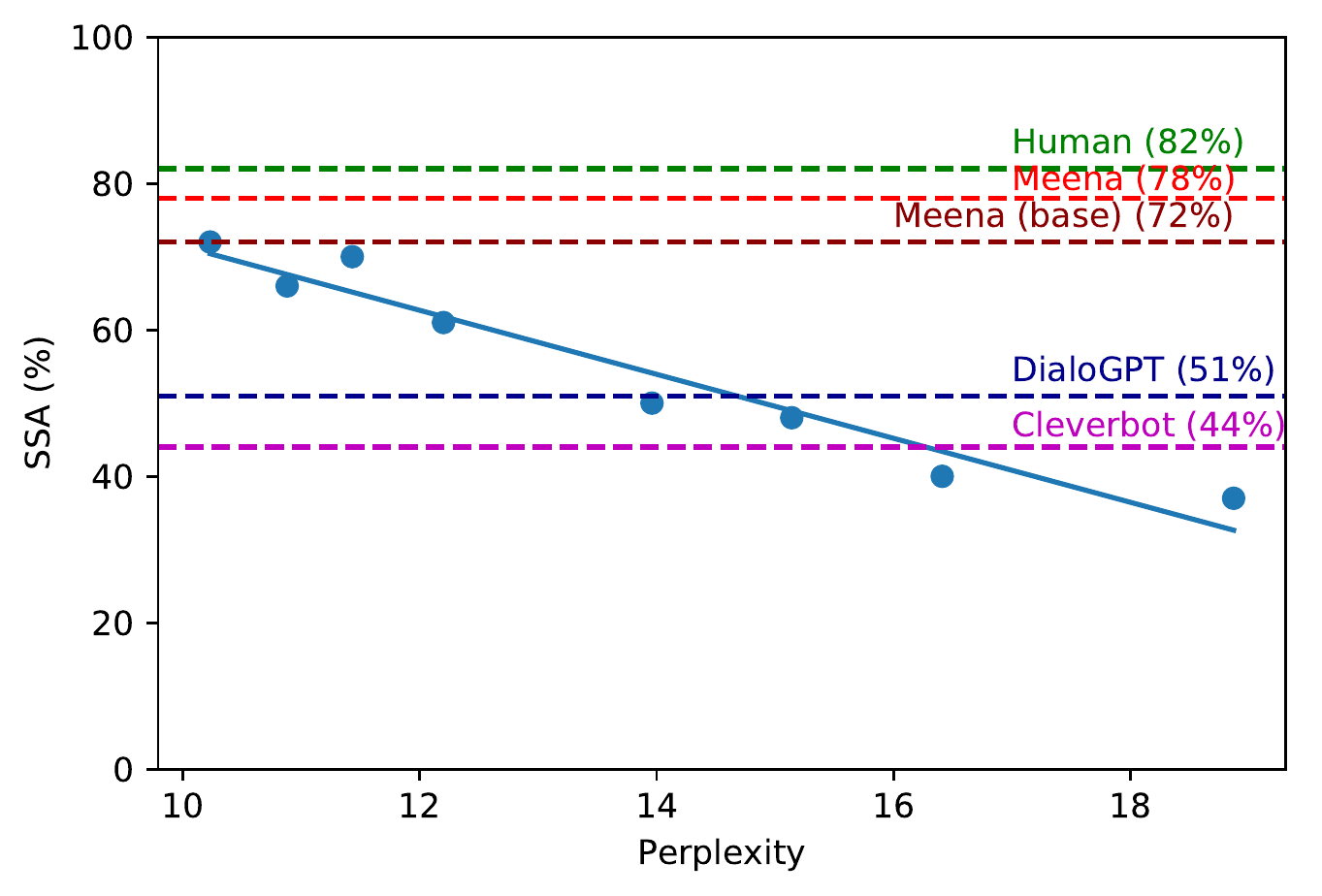}
    \end{subfigure}
    \hfill
    \begin{subfigure}[t]{0.32\textwidth}
        \includegraphics[width=1.0\textwidth]{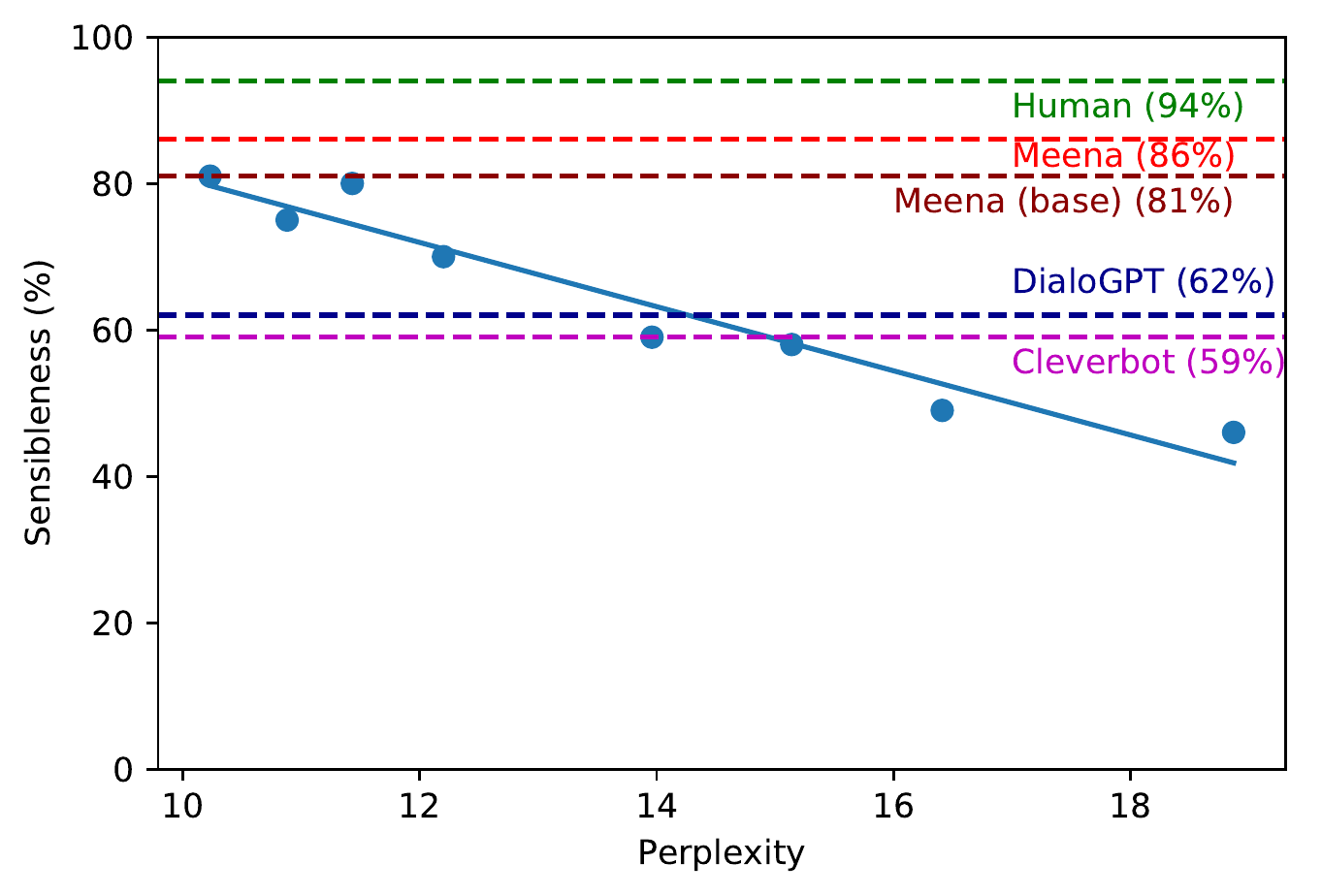}
    \end{subfigure}
    \hfill
    \begin{subfigure}[t]{0.32\textwidth}
        \includegraphics[width=1.0\textwidth]{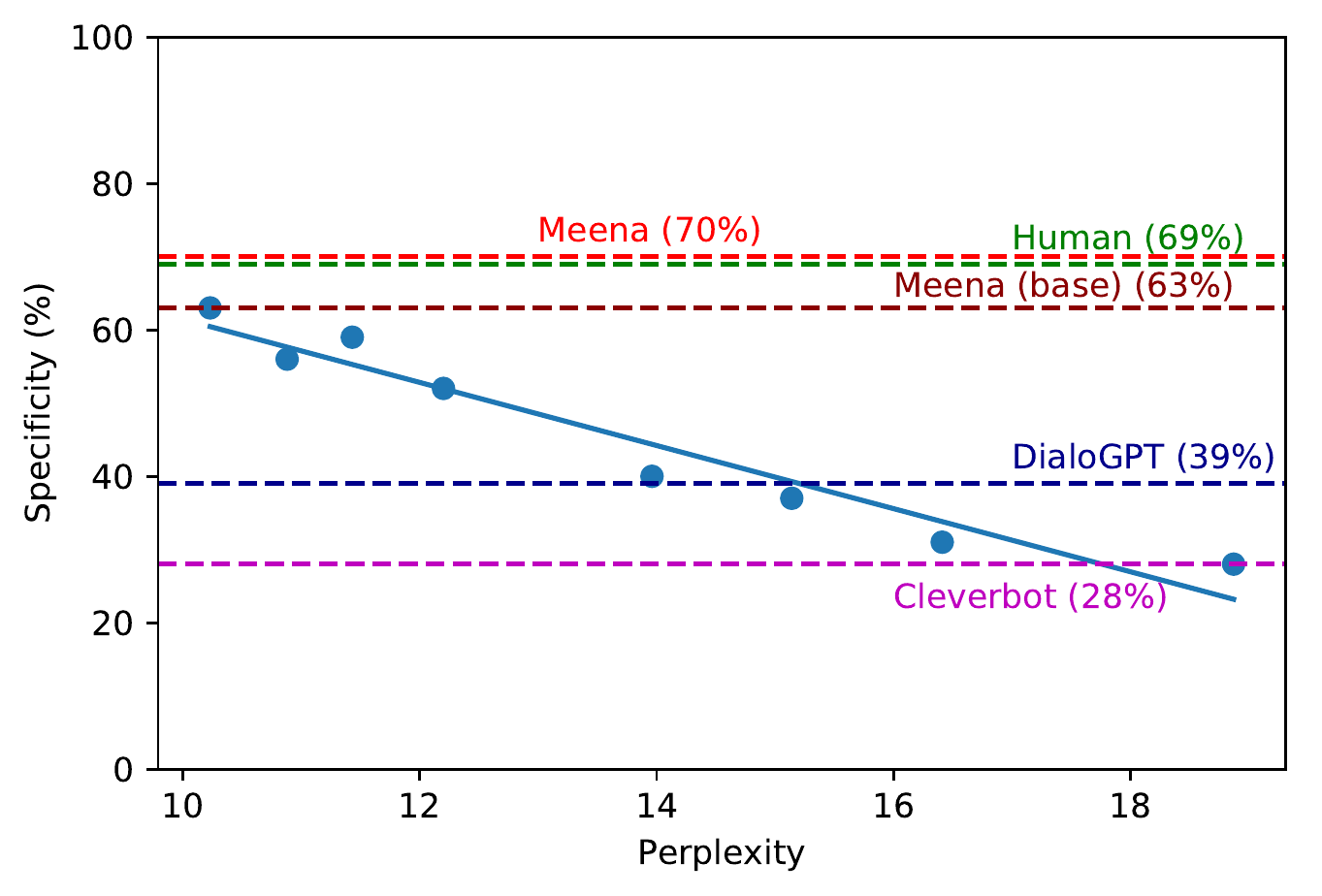}
    \end{subfigure}
    \caption{Static evaluation results.\label{fig:static-evaluation}}    
\end{figure*}

The correlation was $R^2=0.93$ for static sensibleness vs perplexity and $R^2=0.94$ for static specificity vs perplexity indicating this might be a good automatic metric for measuring sensibleness and specificity. Static SSA vs perplexity has $R^2=0.94$. The static evaluation results are shown in Figure \ref{fig:static-evaluation}. The correlation is close to linear, but it is unclear whether the trend will continue for even lower values of perplexity.

In interactive evaluation (Section \ref{subsec:interactive}) crowd workers could chat about anything they wanted. We observe similarly strong correlation with perplexity (see \Cref{fig:ppl-isensibleness,fig:ppl-ispecificity,fig:ppl-issa}) and very similar sensibleness and specificity values as the static evaluation. This indicates that the static evaluation correlation with perplexity is not due to dataset bias.

Regarding consistency, the lowest perplexity model was evaluated 7 times with static evaluations and also 7 times with interactive evaluations. Each time, we obtained a different set of randomly sampled responses. Across the evaluations the standard deviation is $2\%$ for static SSA and is $1\%$ for interactive SSA, indicating that both metrics are consistent enough for our purposes.

\subsection{Human-level Estimates}
\label{sec:human_level}
As expected, human sensibleness is very high, but it is not perfect. Human sensibleness was estimated at 94\% static and 97\% interactive. People have misunderstandings, miss attempts at humor and sometimes lack shared context or background. Also aligned with intuition, humans are sometimes not specific due to momentary lack of ideas, interest or knowledge. The human specificity scores are 69\% static and 75\% interactive. The resulting SSAs are 82\% static and 86\% interactive.

\begin{table*}[tbh!]
  \centering
  \resizebox{16cm}{!}{
  \begin{tabular}{p{6cm}|p{4cm}|p{4cm}|p{4cm}}
  \toprule
{\bf Sample Contexts} & {\bf Meena (base)} & {\bf Cleverbot} & {\bf DialoGPT} \\
\midrule
% 3-turn contexts
\aturn{You must have read a lot of books} \newline \bturn{Oh wow, how do you know? Do I look nerdy?} & \bturn{So how do you know?} & \bturn{Never. said I was.} & \bturn{Oh thank goodness.}  \\
\aturn{Of course, you're not!} & \rates{sensible: 1, specific: 1} & \rates{sensible: 0, specific: 0}  & \rates{sensible: 1, specific: 1} \\
\cdashlinelr{1-4}
\aturn{You know nothing, Jon Snow!} \newline \bturn{You must be a fan of Game of Thrones}  & \bturn{Who's your favorite character?} & \bturn{What do you mean?} & \bturn{You can't. But that's not why you shouldn't know.} \\
\aturn{Haha yea, how can I not?} & \rates{sensible: 1, specific: 1} & \rates{sensible: 0, specific: 0} & \rates{sensible: 0, specific: 0} \\
\cdashlinelr{1-4}
\aturn{What is your favorite island in the world?} \newline \bturn{I would say Fiji} & \bturn{Just a beautiful place} & \bturn{The title is self explanatory.} & \bturn{You can see some great scenery.} \\
\aturn{Interesting, what's nice about it?} & \rates{sensible: 1, specific: 0} & \rates{sensible: 0, specific: 0} & \rates{sensible: 1, specific: 0} \\
\midrule
% 2-turn contexts
\multirow{2}{6cm}{\aturn{You must have read a lot of books} \newline \bturn{Oh wow, how do you know? Do I look nerdy?}} &  \aturn{I know because I've read a lot of books.} & \aturn{Observation.} & \aturn{I know because you look like your mom.} \\
& \rates{sensible: 1, specific: 1} & \rates{sensible: 1, specific: 0} & \rates{sensible: 1, specific: 1} \\
\cdashlinelr{1-4}
\multirow{2}{6cm}{\aturn{You know nothing, Jon Snow!} \newline \bturn{You must be a fan of Game of Thrones}}  & \aturn{Yeah, I'm reading the books at the moment.}  & \aturn{No.} & \aturn{I've never really been interested in it...} \\
& \rates{sensible: 1, specific: 1} & \rates{sensible: 1, specific: 0} & \rates{sensible: 1, specific: 1} \\
\cdashlinelr{1-4}
\multirow{2}{6cm}{\aturn{What is your favorite island in the world?} \newline \bturn{I would say Fiji}} & \aturn{Fiji, yes! I've been there once. It's the most beautiful place I've ever been. It definitely makes me want to move there. I hope you get to visit again soon!} & \aturn{I would have guessed 666.} & \aturn{Fiji is awesome! But I would say Fiji and Oceania are more popular at the moment.} \\
& \rates{sensible: 1, specific: 1} & \rates{sensible: 0, specific: 0}  & \rates{sensible: 0, specific: 0} \\
\midrule
% 1-turn
\multirow{2}{6cm}{\aturn{You must have read a lot of books}} & \bturn{Reading's for nerds} & \bturn{What books have you read.} & \bturn{I read a lot of books but have never read a single book about the history of our planet.} \\
& \rates{sensible: 1, specific: 1} & \rates{sensible: 1, specific: 1} & \rates{sensible: 1, specific: 1} \\
\cdashlinelr{1-4}
\multirow{2}{6cm}{\aturn{You know nothing, Jon Snow!}} & \bturn{Thank you for this. I was hoping I wasn't the only one getting an ASOIAF vibe from this.} & \bturn{Am not jon snow.} & \bturn{The North Remembers...} \\
& \rates{sensible: 1, specific: 1} & \rates{sensible: 1, specific: 0} & \rates{sensible: 1, specific: 1} \\
\cdashlinelr{1-4}
\multirow{2}{6cm}{\aturn{What is your favorite island in the world?}} & \bturn{That's a tough one. I'd have to say Hawaii.} & \bturn{Daniel I like frites.} & \bturn{Probably the one on the north island} \\
& \rates{sensible: 1, specific: 1} & \rates{sensible: 0, specific: 0} & \rates{sensible: 0, specific: 0} \\
\bottomrule
  \end{tabular}
  }
  \caption{{\bf Sample responses from static evaluation} -- Shown are responses of Meena (base), Cleverbot, and DialoGPT on some sample contexts together with their binary labels per category (sensibleness and specificity). Note that the contexts were constructed in advance before we generate all model responses. These contexts are not in the MTB dataset.}
  \label{t:static_analysis}
\end{table*}

\subsection{XiaoIce, Mitsuku, DialoGPT and Cleverbot}
\label{sec:results_other_bots}
Crowd workers labeled 1,173 XiaoIce turns within their original conversation context. Per these labels, XiaoIce scores 31\% interactive SSA which is comprised of 45\% sensibleness and 17\% specificity. We used majority voting of 5 workers per chatbot response. Agreement between workers was 77\% for sensibleness and 81\% for specificity and Krippendorff's alpha was 0.54 for sensibleness and 0.40 for specificity (which indicates fairly strong agreement). For further verification of the results, we also had a group of 4 internal company volunteers that are native Mandarin speakers to label a subset of 25 conversations (~247 chatbot turns). The volunteers did not know the crowd worker results. The volunteer based estimate is 36\% interactive SSA with 53\% sensibleness and 19\% specificity.

Volunteers report that XiaoIce responses display empathy and personality. However, the responses do not always make sense and are often generic. The generic responses exuded a fun personality, but were not necessarily related to the ongoing conversation. As mentioned above, XiaoIce is reported to be optimized for long term engagement \citep{xiaoice}. It is also designed with a multifaceted notion of `humanlikeness' in mind, emphasising interpersonal qualities such as empathy. As we discuss in Section~\ref{sec:discussion}, a broader metric of humanlikeness would be a useful topic for future research.

\citet{see19} report engagingness is not the same as humanness. In that work, they control and optimize certain attributes of response generation. One of the results is that their chatbots achieve near-human-level engagingness, but simultaneously are not very close to human-level humanness. This along with limitations described in Section \ref{sec:xiaoice_mitsuku_eval} may help explain why XiaoIce's sensibleness and specificity scores seem relatively low in our evaluation.

Mitsuku scores 56\% interactive SSA with 72\% sensibleness and 40\% specificity over 899 chatbot turns. Mitsuku has won the Loebner prize 5 times,\footnote{\url{https://en.wikipedia.org/w/index.php?title=Loebner_Prize&oldid=927503336}} however, we learned from the Pandorabots team that the website version we tested with was not the one that competed in the Loebner prize. Optimizing human likeness for the competition might correlate more with our metrics.

Using crowd workers, we executed interactive evaluation of DialoGPT over 300 conversations (2,660 chatbot turns labeled). DialoGPT scores $48\% \pm 3\%$ interactive SSA with $57\% \pm 2\%$ sensibleness and $39\% \pm 4\%$ specificity. It obtained similar results in static evaluation. In its paper, DialoGPT attains performance close to human in a single-turn setting. On the other hand, our human evaluation is multi-turn, which is expected to be more challenging. Additionally, DialoGPT scores poorly on specificity, and our impression from browsing transcripts is that it prefers briefer and more generic responses. This might be because the model is optimized for classic Turing-test evaluation, in which overly chatty responses increase the risk of making a mistake. These results and conjectures come with the caveat, as described above, that we wrote our own decoder for this model since the public DialoGPT codebase does not yet have one.

Cleverbot, unlike Meena and DialoGPT, performs notably better on interactive rather than static evaluation. It scores interactive SSA 56\% and static SSA 44\%. Interactive specificity, 45\%, is especially higher than its static counterpart, 28\%. Upon closer inspection of the data, we hypothesize that: (1) in the interactive setting, Cleverbot has opportunities to steer the conversation towards topics that it is more familiar with; (2) the minimum interactive conversation length of 14 turns makes it possible for a significant portion of these turns to be greetings and goodbyes, which both Cleverbot and Mitsuku are consistent in appropriately responding to. Furthermore, the interactive SSA scores for Mitsuku and Cleverbot are the same, 56\% when averaging sensibleness and specificity before rounding. Mitsuku scores higher sensibleness (72\% versus 68\%), but lower specificity (40\% versus 45\%). It seems that relative to Mitsuku, Cleverbot replies more often in ways that are borderline nonsensical and lack consistent personality. Finally, we remark that the standard deviation of the Cleverbot interactive SSA is $\pm1\%$ across two interactive evaluation sessions.\footnote{Due to technical issues when calling the Cleverbot API we only collected 195 interactive conversations (1,751 chatbot turns labeled) instead of the 300 conversations which we collected for DialoGPT.}

\subsection{Sample Responses: \meenabase{}, Cleverbot, and DialoGPT}
To understand how Meena qualitatively compares to other models, we show in Table~\ref{t:static_analysis} sample responses from \meenabase{}, Cleverbot, and DialoGPT under the same set of contexts (which were constructed before we generate all model responses). For 1- and 2-turn contexts,  responses from Meena base are all sensible and specific.
In addition, \meenabase{} generates rich and interesting responses, e.g., the mention of \textit{``ASOIAF vibe''} to refer to ``A Song of Ice and Fire'' in the famous Game of Thrones series or the remark about Fiji island being \textit{``the most beautiful place I've ever been''}.

In contrast, Cleverbot can generate sensible responses for some contexts, but they are not always specific, e.g., Cleverbot replied with \textit{``Observation''} and \textit{``No''}. DialoGPT is more specific and can also generate interesting responses, e.g., \textit{``The North Remembers ...'''}. However, it does not make sense at times, e.g., in-turn contradiction in this response \textit{``Fiji is awesome! But I would say Fiji and Oceania are more popular ...''} or vague answer \textit{``Probably the one on the north island''}.

When it comes to longer (3-turn) contexts in Table~\ref{t:static_analysis}, \meenabase{} continues to generate high-quality responses, whereas none of Cleverbot's responses are sensible.
DialoGPT is more sensible and specific than Cleverbot, but less so than \meenabase{}.

\section{Further Advancing SSA}
\label{sec:advance_ssa}
In this section we take the interactive SSA from $72\% \pm 1\%$, for \meenabase{}, to $79\% \pm 1\%$, for \meenafull{}, by further tuning our decoding strategy and adding a rule to detect cross turn repetitions.

\subsection{Advancing Decoding}

We evaluate both temperature $T$ and top-\textit{k} to mitigate negative effects from the tail of the distribution \citep{holtzman19}. We chose top-\textit{k} (\textit{k} = 40) and $T=1.0$ following \citet{fan18, gpt2, keskar19, ippolito19b}. With this setting and maintaining $N=20$, we note an SSA increase from 72\% to 74\% relative to sampling from the whole vocabulary with $T=0.88$. This result is the same for both the interactive and the static evaluation.

\begin{figure}[ht!]
\centering
\includegraphics[width=230pt]{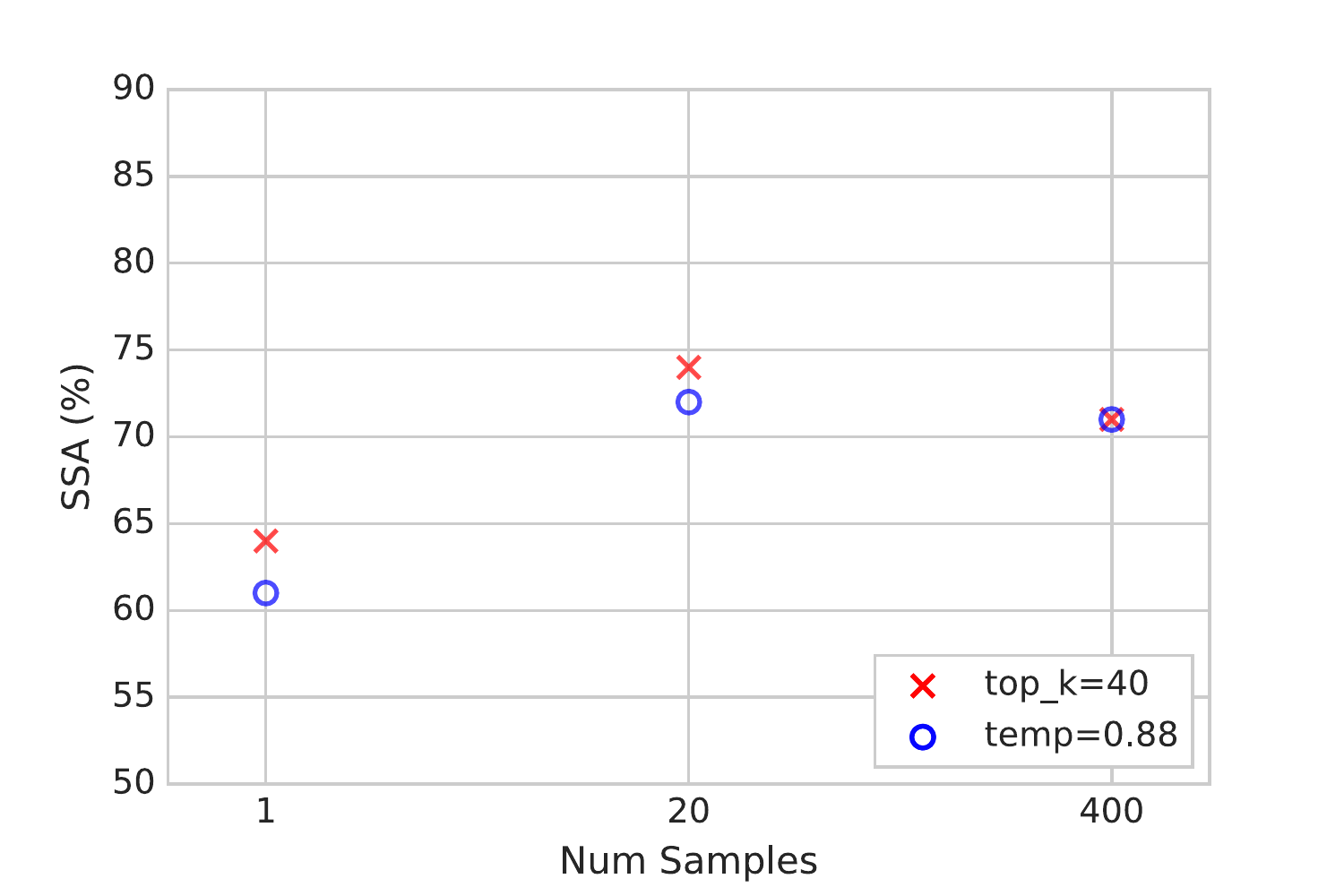}
\caption{Static SSA over number of sampled responses for top-\textit{k} and sampling with temperature.}
\label{fig:decoding-ssa}
\end{figure}

We also swept the number of samples in sample-and-rank, evaluating $N \in \{1, 20, 400\}$. The results show that $N=20$ provides a significant improvement over $N=1$, with an absolute improvement in SSA of ${\sim}10\%$ (Figure \ref{fig:decoding-ssa}). However, $N=400$ demonstrates worse performance for sensibleness (Figure \ref{fig:decoding-sensibleness} in the appendix) and diminishing returns over $N=20$ for specificity (Figure \ref{fig:decoding-specificity} in the appendix). The significant improvement from sample-and-rank with $N=20$ motivates future work exploring alternate ranking functions and tuning parameters.

\subsection{Addressing Cross-turn Repetitions}

In interactive evaluation, about one third of the conversations with \meenabase{} contain cross-turn repetitions toward the end. Cross-turn repetition means that one turn somewhat repeats an earlier turn. For illustration, we cherry picked particularly problematic examples of cross-turn repetition shown in Tables \ref{t:rep1} and \ref{t:rep2}.

\begin{table}
\small
\centering
\begin{tabular}{rp{6cm}}
\toprule
\multicolumn{2}{c}{{\bf Cross-turn Repetition Example 1}} \\
\midrule
\textbf{Human}: & \textit{Hi!} \\
\textbf{Meena}: & \textit{Hi! How are you?! :)} \\
\textbf{Human}: & \textit{doing good, what would be the best city to visit in Europe?} \\
\textbf{Meena}: & \textit{Paris, Barcelona, Amsterdam, Prague} \\
\textbf{Human}: & \textit{great list! why?} \\
\textbf{Meena}: & \textit{Paris is a beautiful city, and Barcelona is too. And I've always wanted to go to Amsterdam and Prague and have never been.} \\
\textbf{Human}: & \textit{have you even been to India?} \\
\textbf{Meena}: & \textit{No, but I'd love to go.} \\
\textbf{Human}: & \textit{how about Japan?} \\
\textbf{Meena}: & \textit{I'd love to go to Japan too.} \\
\textbf{Human}: & \textit{how about Paris?} \\
\textbf{Meena}: & \textit{I'd love to go to Paris too.} \\
\textbf{Human}: & \textit{are you lying to me?} \\
\textbf{Meena}: & \textit{I'd love to go to Japan too.} \\
\textbf{Human}: & \textit{i think you're lying to me, that makes me sad} \\
\textbf{Meena}: & \textit{I'd love to go to Japan, too.} \\
\textbf{Human}: & \textit{you're blocking me out! you have to talk to me} \\
\textbf{Meena}: & \textit{I'd love to go to Japan.} \\
\textbf{Human}: & \textit{where else?} \\
\textbf{Meena}: & \textit{I'd love to go to Japan, too!} \\
\bottomrule
\end{tabular}
\caption{Multiple near identical repetitions of \textit{``I'd love to go to Japan too."}.\label{t:rep1}}
\end{table}

\begin{table}
\small
\centering
\begin{tabular}{rp{6cm}}
\toprule
\multicolumn{2}{c}{{\bf Cross-turn Repetition Example 2}} \\
\midrule
\textbf{Human}: & \textit{Hi!} \\
\textbf{Meena}: & \textit{Hey there! How are you doing?} \\
\textbf{Human}: & \textit{not well, i'm quite depressed} \\
\textbf{Meena}: & \textit{That's no good. Wanna talk about it?} \\
\textbf{Human}: & \textit{not really, it'll only make it worse...can you cheer me up?} \\
\textbf{Meena}: & \textit{Sure! What's got you down?} \\
\textbf{Human}: & \textit{I'd love to hear a really lame joke} \\
\textbf{Meena}: & \textit{How about this one? What did the farmer say when he lost his tractor? Where is my tractor?} \\
\textbf{Human}: & \textit{haha that was lame! where did you hear it?} \\
\textbf{Meena}: & \textit{I was at the mall and I heard it in the food court. How was your day?} \\
\textbf{Human}: & \textit{wow, i didn't know you could walk!} \\
\textbf{Meena}: & \textit{Why did the scarecrow win an award?} \\
\textbf{Human}: & \textit{i don't know..you tell me!} \\
\textbf{Meena}: & \textit{Because he was outstanding in his field!} \\
\textbf{Human}: & \textit{haha, where did you hear that?} \\
\textbf{Meena}: & \textit{I was in the food court and I heard it in the field!} \\
\textbf{Human}: & \textit{you have a great sense of humor!} \\
\textbf{Meena}: & \textit{Why did the chicken cross the road? He was outstanding in his field!} \\
\textbf{Human}: & \textit{ok, i don't want to hear any more jokes} \\
\textbf{Meena}: & \textit{Do you know why he was outstanding in his field? Because he was a chicken!} \\
\textbf{Human}: & \textit{haha looks like you're really into this discussion!} \\
\textbf{Meena}: & \textit{Yes! I am! Do you know why? Because he was outstanding in his field!} \\
\textbf{Human}: & \textit{haha, borderline creepy but still funny :D} \\
\textbf{Meena}: & \textit{Why was the mouse trapped in the bag? Because he was outstanding in his field!} \\
\textbf{Human}: & \textit{i think i broke you} \\
\textbf{Meena}: & \textit{Do you know why I broke you? Because you were outstanding in your field!} \\
\bottomrule
\end{tabular}
\caption{Multiple partial repetitions of \textit{``(...) Because he was outstanding in his field!"}.\label{t:rep2}}
\end{table}

It is worth mentioning that there also exist in-turn contradictions and repetitions, where the contradiction or repetition is contained in the response turn itself (e.g., \textit{``I like pizza, but I don't like it"}). This type of artifact is often observed in Meena versions with worse perplexities, but is far less frequent in the \meenabase{}, which has the lowest perplexity as reflected in the samples shared in the appendix and the higher sensibleness scores.

We wrote a rule that detects if any two turns contain long common sub-sequences. We automatically remove candidates that are detected as repetition. This rule seems to have addressed most of the cross-turn repetition. We, therefore, further improve on the above interactive SSA of $74\% \pm \%1$ to $79\% \pm 1\%$.

\subsection{Safety Layer}
It is important to mention that the evaluation and conversation collection for the full Meena version was conducted with an additional classifier layer at serving time as part of the filtering mechanism to automatically filter out potentially sensitive or toxic response candidates for publication.

\section{Related Work}
\label{sec:related}
Finding a good automatic metric that correlates with human evaluation has been an important goal of open-domain conversational modeling. BLEU~\cite{Papineni02bleu}, ROUGE~\cite{rouge}, or other related metrics in translation and summarization, while popular and easy to compute, have been shown to be unsuitable for dialog \citep{liu16} or more broadly language generation systems \cite{novikova17}.

Past works have attempted to build learnable metrics, either in a supervised fashion~\cite{lowe17}, which requires human labels, or with unsupervised approaches~\cite{ruber,ghazarian19}, that are more complex and need separate training, e.g., of a ranking system. In our work, we show that perplexity, which is readily available to any neural seq2seq model, exhibits a strong correlation with human evaluation.
Our work is therefore also related to past attempts to correlate perplexity with other automatic metrics
in other tasks, e.g., perplexity vs.~BLEU in translation~\cite{luong15}.

Another interesting line of work is to combine human evaluation with either automatic metrics~\cite{chaganty18} or with model likelihood~\cite{hashimoto19}. While theoretically motivated, these metrics are too complex to be practical, requiring both human judgments and training separate models, e.g., an estimator~\cite{chaganty18} to reduce bias in automatic evaluation or a discriminator~\cite{hashimoto19} to distinguish between human- and model-generated samples.

In terms of designing of human evaluation metrics, existing literature differs in what attributes are used to assess the quality of a neural conversational model. Many works, e.g., \citet{zhao17,xu18,ippolito19}, have focused solely on the diversity aspect to counter the commonly observed problem of models generating generic responses \cite{li16diversity}.
Others have attempted to improve and evaluate multiple aspects at once. For example, 
\citet{amazoneval} aim to unify many metrics, such as diversity, engagement, and user experience; \citet{gao19joint} jointly optimize for both diversity and relevance; \citet{see19} control decoding attributes (such as repetition, specificity, response-relatedness, and question-asking) to improve engagingness and interestingness; and \citet{hashimoto19} design metrics to capture human likeness and diversity.

In contrast, we focus on sensibleness and specificity for our human evaluation. 
While human likeness and relevance used in aforementioned works are related to sensibleness, we specifically use sensibleness as it leads to better agreement among crowd workers (see $\S$\ref{subsec:metrics}). 
Similar reasoning applies to specificity, which is related to other attributes such as engagingness and interestingness, as measured in previous works.\footnote{It is worth pointing out that we do not explicitly measure diversity as it requires judging a set of responses; whereas, for conversation, what is most important is the first reply that a chatbot produces.
As our decoding method is sampling, it implies that our generation is diverse. However, there remains a question of whether the sampled response is of high quality. The fact that our model has low perplexity and achieves high SSA score indicates that the generation is meaningful.
}
A limitation of our work is that it does not cover aspects such as empathy \cite{xiaoice,rashkin18}.

While we do not explicitly control for specificity, existing works, such as \cite{zhang18control,ko19}, attempted to do so by augmenting the decoder of seq2seq models with specificity-control components. These added complexities sometimes lead to implausible responses as analyzed by \citet{ko19}. In contrast, the specificity of our model improves as perplexity decreases.

Recent work on DialoGPT \cite{dialogpt} compares the conversation quality of chatbots with that of humans but their evaluation settings are limited to single-turn dialogs. We instead conduct our evaluation on conversations of up to 3 turns in the static MTB benchmark and 14 turns in the interactive setup.

\section{Discussion}
\label{sec:discussion}
Our results suggest perplexity on public domain social media conversations might be a good automatic proxy for human judgement of fundamental attributes of human-likeness, such as sensibleness and specificity. The results also suggests that optimizing the probability of the next token on larger volumes of social media conversations could lead to human-like sensibleness in an open-domain setting. However, our static evaluation dataset only contains one to three-turn contexts and is biased by the sources of the first turn and the fact that the two-turn and three-turn contexts build on the shorter contexts. Moreover the contexts in this dataset are predominantly Turing test and social conversation style, including common sense, basic knowledge, asking/sharing about personality, likes/dislikes, opinions, feelings, hobbies, pleasantries, etc. This dataset does not include contexts like deeper question answering (e.g., how fast is a cheetah), basic math (e.g., how much is 1+1) and common sense tests designed to challenge machines, but not humans \cite{levesque11}. Human-likeness is an incredibly broad and abstract concept. The interactive evaluation addresses some of the bias and scope limitations in static evaluation while still providing a consistent score to quantify a given chatbot. Nevertheless, unlike static evaluation it does not allow for granular comparison between different chatbot responses. In addition, it may be too short (14 to 28 turns), and may assign too much weight to typical beginning and ending of conversations. It may also be too short to cover deeper topics and exercise longer term memory.

Furthermore, it may be necessary to expand the set of basic human-like conversation attributes being measured beyond sensibleness and specificity. Some directions could include humor, empathy, deep reasoning, question answering and knowledge discussion skills. One could also break down sensibleness into its implicit sub-components: logical and personality consistency, common sense, relevance, basic factual correctness and so on. Future work may also explore the continued optimization of sensibleness via the optimization of test set perplexity.

\section*{Acknowledgments}
Thanks to the people who gave feedback on drafts of the paper: Anna Goldie, Abigail See, Yizhe Zhang, Lauren Kunze, Steve Worswick, Jianfeng Gao, Daphne Ippolito, Scott Roy, Ilya Sutskever, Tatsu Hashimoto, Dan Jurafsky, Dilek Hakkani-tur, Noam Shazeer, Gabriel Bender, Prajit Ramachandran, Rami Al-Rfou, Michael Fink, Mingxing Tan, Maarten Bosma and Adams Yu. Also thanks to the many volunteers who helped collect conversations with each other and with various chatbots. Finally thanks to Samy Bengio, Noam Shazeer, Anna Goldie, Rami Al-Rfou, Khoa Vo, Trieu H. Trinh, Ni Yan, Kyu Jin Hwang and the Google Brain team for the help with the project. \\

\bibliography{meena}
\bibliographystyle{acl_natbib}

\clearpage
\appendix
\onecolumn
\twocolumn
\section{Additional Sample Conversations}
\label{apdx:conversations}
With the help of many internal company volunteers we collected a total of about 100 conversations with Mitsuku, XiaoIce and \meenafull{}. The conversations are available on Github \footnote{\url{https://github.com/google-research/google-research/tree/master/meena/}}. This section contains samples obtained by random shuffling these sets, and taking the first 10. Conversations were collected following the standard instructions for interactive evaluation where the human starts. Therefore, conversations are supposed to start with ``Hi!", contain between 16 and 32 turns in total, and are open-domain with no particular topic. Nevertheless, some participants did not follow the first-turn rule strictly, so some conversations may start with for instance ``hi there" instead of ``Hi!". Also, a few conversations are under or over the length limits.

Unlike in Section \ref{sec:sample_conversations}, which contains cherry picked samples, we present random samples of everything that was collected after a few potentially sensitive conversations have been removed from the original sets. We also redacted potential personally identifiable information and indicated that with the word ``REDACTED". Finally, please note that both XiaoIce and Mitsuku sometimes include an image in their reply and occasionally, volunteers include text descriptions of the images they see.

\subsection{Meena}
The following are a sample of the conversations with the \meenafull{} ($79\% \pm 1\%$ interactive SSA).

\FloatBarrier
\begin{table}[!htbp]
\small
\centering
\begin{tabular}{rp{6cm}}
\toprule
\multicolumn{2}{c}{{\bf Meena Conversation 1}} \\
\midrule
\textbf{Human}: & \textit{Hi!} \\
\textbf{Meena}: & \textit{Hey there! What's up?} \\
\textbf{Human}: & \textit{Same old, Monday blues...what are you up to?} \\
\textbf{Meena}: & \textit{I'm procrastinating on a paper I'm supposed to be writing.} \\
\textbf{Human}: & \textit{are you thinking of pulling an all nighter?} \\
\textbf{Meena}: & \textit{Possibly. What's your paper on?} \\
\textbf{Human}: & \textit{Nah, i just do the grunge work around here...one day I wish to write a paper that will change the world!} \\
\textbf{Meena}: & \textit{What's the topic?} \\
\textbf{Human}: & \textit{not sure, any suggestions?} \\
\textbf{Meena}: & \textit{Well what's the assignment on?} \\
\textbf{Human}: & \textit{breathing under water} \\
\textbf{Meena}: & \textit{How to breathe underwater.} \\
\textbf{Human}: & \textit{that's what i'm hoping to figure out!} \\
\textbf{Meena}: & \textit{I hope you figure it out! Because I can't.} \\
\textbf{Human}: & \textit{i'm sure you can if you try :) you're the best!} \\
\textbf{Meena}: & \textit{Thank you, but I'm pretty sure you're the best.} \\
\textbf{Human}: & \textit{how was your weekend?} \\
\textbf{Meena}: & \textit{Pretty good. Yours?} \\
\textbf{Human}: & \textit{good, I practiced the violin} \\
\textbf{Meena}: & \textit{That's awesome! How long have you been playing?} \\
\textbf{Human}: & \textit{just started learning last month} \\
\textbf{Meena}: & \textit{That's awesome! What made you want to learn the violin?} \\
\bottomrule
\end{tabular}
\end{table}

\begin{table}[!htbp]
\small
\centering
\begin{tabular}{rp{6cm}}
\toprule
\multicolumn{2}{c}{{\bf Meena Conversation 2}} \\
\midrule
\textbf{Human}: & \textit{Hi!} \\
\textbf{Meena}: & \textit{Hello! How are you? :)} \\
\textbf{Human}: & \textit{I'm doing okay. How about you?} \\
\textbf{Meena}: & \textit{I'm doing pretty good, thanks for asking! :)} \\
\textbf{Human}: & \textit{What's your name?} \\
\textbf{Meena}: & \textit{It's Kimberly! What's your name?} \\
\textbf{Human}: & \textit{That's a nice name. My name is Paul.} \\
\textbf{Meena}: & \textit{Nice to meet you Paul! :)} \\
\textbf{Human}: & \textit{Nice to meet you as well Kimberly} \\
\textbf{Meena}: & \textit{So, what do you like to do?} \\
\textbf{Human}: & \textit{I like going to the movies} \\
\textbf{Meena}: & \textit{Cool! What's your favorite movie?} \\
\textbf{Human}: & \textit{Pulp Fiction} \\
\textbf{Meena}: & \textit{Nice! I love Reservoir Dogs. :)} \\
\textbf{Human}: & \textit{Yes, that movie is good as well} \\
\textbf{Meena}: & \textit{Who's your favorite character in Reservoir Dogs?} \\
\bottomrule
\end{tabular}
\end{table}

\begin{table}[H]
\small
\centering
\begin{tabular}{rp{6cm}}
\toprule
\multicolumn{2}{c}{{\bf Meena Conversation 3}} \\
\midrule
\textbf{Human}: & \textit{Hi!} \\
\textbf{Meena}: & \textit{Hello! What are you up to tonight?} \\
\textbf{Human}: & \textit{nothing much, why?} \\
\textbf{Meena}: & \textit{Just wondering what you are up to!} \\
\textbf{Human}: & \textit{getting something to eat and then watch something} \\
\textbf{Meena}: & \textit{Whatcha gonna eat?} \\
\textbf{Human}: & \textit{no idea yet, do you have a recommendation on what to watch on Netflix?} \\
\textbf{Meena}: & \textit{What kind of movies do you enjoy?} \\
\textbf{Human}: & \textit{I'm ok with most things, if you know a good movie, I'll give it a try} \\
\textbf{Meena}: & \textit{Have you seen 12 Angry Men or In Bruges?} \\
\textbf{Human}: & \textit{no and no} \\
\textbf{Meena}: & \textit{I highly recommend them, they are both great movies.} \\
\textbf{Human}: & \textit{what genre of movie are they in?} \\
\textbf{Meena}: & \textit{Both are dramas} \\
\textbf{Human}: & \textit{are they american dramas?} \\
\textbf{Meena}: & \textit{Yes, both are American} \\
\textbf{Human}: & \textit{ok, who are the main characters?} \\
\textbf{Meena}: & \textit{Kristen Stewart and Ben Affleck} \\
\textbf{Human}: & \textit{in which movie?} \\
\textbf{Meena}: & \textit{The Runaways} \\
\textbf{Human}: & \textit{so not in any of the two you mentioned earlier?} \\
\textbf{Meena}: & \textit{No, they aren't in the movies I mentioned earlier} \\
\textbf{Human}: & \textit{ok, sounds good} \\
\textbf{Meena}: & \textit{So, you haven't seen them?} \\
\textbf{Human}: & \textit{no, none of them} \\
\textbf{Meena}: & \textit{Which ones have you seen?} \\
\textbf{Human}: & \textit{none of them} \\
\textbf{Meena}: & \textit{Not even Spider-Man 1?} \\
\textbf{Human}: & \textit{no} \\
\textbf{Meena}: & \textit{That's a shame. It's a pretty good movie.} \\
\bottomrule
\end{tabular}
\end{table}

\vspace{4mm}
\begin{table}[!hbtp]
\small
\centering
\begin{tabular}{rp{6cm}}
\toprule
\multicolumn{2}{c}{{\bf Meena Conversation 4}} \\
\midrule
\textbf{Human}: & \textit{Hi!} \\
\textbf{Meena}: & \textit{Hey! How's it going?} \\
\textbf{Human}: & \textit{How's what going?} \\
\textbf{Meena}: & \textit{Pretty good. How's your day been?} \\
\textbf{Human}: & \textit{Excellent. I spent most of trying to learn to surf.} \\
\textbf{Meena}: & \textit{Sweet! What kind of surfing?} \\
\textbf{Human}: & \textit{The kind where I try to catch a wave and fail} \\
\textbf{Meena}: & \textit{Haha. Do you have a surfboard?} \\
\textbf{Human}: & \textit{I borrowed a friend's, to see if I like it. I'm still a beginner.} \\
\textbf{Meena}: & \textit{Ah. I'm sure you'll get the hang of it.} \\
\textbf{Human}: & \textit{Appreciate the vote of confidence :) Have you ever been surfing?} \\
\textbf{Meena}: & \textit{I have! It's actually a ton of fun.} \\
\textbf{Human}: & \textit{Did you find it hard at first?} \\
\textbf{Meena}: & \textit{A little bit, yes. But once I got the hang of it, it was really fun.} \\
\textbf{Human}: & \textit{That's good to know. Do you go surfing regularly?} \\
\textbf{Meena}: & \textit{Yup. I live in southern Arizona, so there's plenty of surfing to be had.} \\
\bottomrule
\end{tabular}
\end{table}

\begin{table}[tbh!]
\small
\centering
\begin{tabular}{rp{6cm}}
\toprule
\multicolumn{2}{c}{{\bf Meena Conversation 5}} \\
\midrule
\textbf{Human}: & \textit{Hi!} \\
\textbf{Meena}: & \textit{Hi! How are you today?} \\
\textbf{Human}: & \textit{It's Friday, Friday, Friday :D} \\
\textbf{Meena}: & \textit{YAY! I love Fridays!} \\
\textbf{Human}: & \textit{Do you like Rebecca Black's Friday?} \\
\textbf{Meena}: & \textit{She's okay. I prefer Friday by Rebecca Sugar.} \\
\textbf{Human}: & \textit{okay. How does that song go?} \\
\textbf{Meena}: & \textit{It's a fun fun fun, fun fun song. :)} \\
\textbf{Human}: & \textit{let's sing along} \\
\textbf{Meena}: & \textit{Fun! My hand is a dolphin!} \\
\textbf{Human}: & \textit{And my cheeks are green} \\
\textbf{Meena}: & \textit{And my fins are red} \\
\textbf{Human}: & \textit{And my eyes are blue} \\
\textbf{Meena}: & \textit{I'm a dolphin! I'm a dolphin! I'm a dolphin!} \\
\textbf{Human}: & \textit{haha... cool!} \\
\textbf{Meena}: & \textit{Dolphin power!} \\
\bottomrule
\end{tabular}
\end{table}

\begin{table}[tbh!]
\small
\centering
\begin{tabular}{rp{6cm}}
\toprule
\multicolumn{2}{c}{{\bf Meena Conversation 6}} \\
\midrule
\textbf{Human}: & \textit{Hi!} \\
\textbf{Meena}: & \textit{Hey! How are you doing?} \\
\textbf{Human}: & \textit{I'm fine, thanks!} \\
\textbf{Meena}: & \textit{Well, I'm glad to hear that!} \\
\textbf{Human}: & \textit{Anything that you are excited about lately?} \\
\textbf{Meena}: & \textit{Not really, I haven't had much to be excited about the past year or so. You?} \\
\textbf{Human}: & \textit{yes, I'm excited about Lunar New Year coming!} \\
\textbf{Meena}: & \textit{Cool, what are you going to do?} \\
\textbf{Human}: & \textit{spending time with friends and family, eating traditional food} \\
\textbf{Meena}: & \textit{Nice, sounds fun!} \\
\textbf{Human}: & \textit{yea, tell me something interesting about you} \\
\textbf{Meena}: & \textit{I guess I'm pretty good at drawing!} \\
\textbf{Human}: & \textit{lovely! what can you draw?} \\
\textbf{Meena}: & \textit{I can draw people, I can do portraits, and I can do animal drawings as well.} \\
\textbf{Human}: & \textit{nice. what about scenery?} \\
\textbf{Meena}: & \textit{Not so well, I'm not that good with trees.} \\
\bottomrule
\end{tabular}
\end{table}

\begin{table}[tbh!]
\small
\centering
\begin{tabular}{rp{6cm}}
\toprule
\multicolumn{2}{c}{{\bf Meena Conversation 7}} \\
\midrule
\textbf{Human}: & \textit{Hi!} \\
\textbf{Meena}: & \textit{Hi! How's it going?} \\
\textbf{Human}: & \textit{doing okay, feeling a little bit under the weather} \\
\textbf{Meena}: & \textit{Aw, I hope you feel better soon!} \\
\textbf{Human}: & \textit{thanks! the weather is getting worse} \\
\textbf{Meena}: & \textit{Oh no, I'm sorry. I hope it gets better for you soon.} \\
\textbf{Human}: & \textit{no, it won't get better any time soon. we have to wait until spring} \\
\textbf{Meena}: & \textit{Aww, sorry to hear that.} \\
\textbf{Human}: & \textit{yea, I'm really feeling under the weather} \\
\textbf{Meena}: & \textit{Aw, well that's no good.} \\
\textbf{Human}: & \textit{yea, any suggestion to get over feeling under the weather} \\
\textbf{Meena}: & \textit{I wish I knew. Sorry I can't help.} \\
\textbf{Human}: & \textit{have you ever been feeling under the weather?} \\
\textbf{Meena}: & \textit{Not that I can remember, no.} \\
\textbf{Human}: & \textit{you must be a superman!} \\
\textbf{Meena}: & \textit{Nah, I'm just lucky, I guess.} \\
\bottomrule
\end{tabular}
\end{table}

\begin{table}[!thbp]
\small
\centering
\begin{tabular}{rp{6cm}}
\toprule
\multicolumn{2}{c}{{\bf Meena Conversation 8}} \\
\midrule
\textbf{Human}: & \textit{Hi!} \\
\textbf{Meena}: & \textit{Hi! How are you today?} \\
\textbf{Human}: & \textit{it is hectic} \\
\textbf{Meena}: & \textit{How is it hectic?} \\
\textbf{Human}: & \textit{I have tons of meetings and have a party tonight!} \\
\textbf{Meena}: & \textit{Oh, fun!} \\
\textbf{Human}: & \textit{thanks} \\
\textbf{Meena}: & \textit{How was the party?} \\
\textbf{Human}: & \textit{it is tonight} \\
\textbf{Meena}: & \textit{What kind of party?} \\
\textbf{Human}: & \textit{dancing!} \\
\textbf{Meena}: & \textit{Awesome! What type of dancing?} \\
\textbf{Human}: & \textit{good question. Any kinds of dancing you like. It is quite diverse  :)} \\
\textbf{Meena}: & \textit{What's your least favorite kind of dancing?} \\
\textbf{Human}: & \textit{hmm, not sure. my favorite is tap dance. it is really fun} \\
\textbf{Meena}: & \textit{Favorite tap dance song?} \\
\textbf{Human}: & \textit{hmm, I am not familiar yet. do you have any recommendation?} \\
\textbf{Meena}: & \textit{Not off the top of my head :(} \\
\textbf{Human}: & \textit{no problems} \\
\textbf{Meena}: & \textit{If you were a fruit, what fruit would you be?} \\
\textbf{Human}: & \textit{interesting question, I like berries so i would like to be one of them} \\
\textbf{Meena}: & \textit{Would you consider yourself a member of the strawberry family?} \\
\textbf{Human}: & \textit{it sounds very sweet :)} \\
\textbf{Meena}: & \textit{If I were a dog, what kind of dog would I be?} \\
\bottomrule
\end{tabular}
\end{table}

\begin{table}[!thbp]
\small
\centering
\begin{tabular}{rp{6cm}}
\toprule
\multicolumn{2}{c}{{\bf Meena Conversation 9}} \\
\midrule
\textbf{Human}: & \textit{Hi!} \\
\textbf{Meena}: & \textit{Hello! How are you doing?} \\
\textbf{Human}: & \textit{I'm good. I'm fishing right now} \\
\textbf{Meena}: & \textit{Cool! What are you fishing for?} \\
\textbf{Human}: & \textit{I'm fishing bass} \\
\textbf{Meena}: & \textit{Cool! What do you catch with it?} \\
\textbf{Human}: & \textit{I'm using a walker lure} \\
\textbf{Meena}: & \textit{Cool! Do you eat the fish?} \\
\textbf{Human}: & \textit{Yes. I happen to like eating fish} \\
\textbf{Meena}: & \textit{Cool! What types of fish do you like to eat?} \\
\textbf{Human}: & \textit{Besides bass, I guess catfish and cod} \\
\textbf{Meena}: & \textit{Cool! What do you catch with catfish?} \\
\textbf{Human}: & \textit{haha you don't use catfish as lure. Catfish is a fish!} \\
\textbf{Meena}: & \textit{Oh, haha, I thought you meant you used a lure to catch catfish} \\.
\textbf{Human}: & \textit{oh I did. anyway do you like fishing?} \\
\textbf{Meena}: & \textit{Yeah, I think it's a lot of fun.} \\
\textbf{Human}: & \textit{where do you usually go for that?} \\
\textbf{Meena}: & \textit{I go to the ocean. Where do you usually go?} \\
\bottomrule
\end{tabular}
\end{table}

\begin{table}[!thbp]
\small
\centering
\begin{tabular}{rp{6cm}}
\toprule
\multicolumn{2}{c}{{\bf Meena Conversation 10}} \\
\midrule
\textbf{Human}: & \textit{Hi!} \\
\textbf{Meena}: & \textit{Hi, how's it going?} \\
\textbf{Human}: & \textit{All good! It's cold} \\
\textbf{Meena}: & \textit{It's pretty cold here too.} \\
\textbf{Human}: & \textit{where do you stay?} \\
\textbf{Meena}: & \textit{Wisconsin. How about you?} \\
\textbf{Human}: & \textit{San Jose, Cali} \\
\textbf{Meena}: & \textit{Oh nice! I've always wanted to go to Cali.} \\
\textbf{Human}: & \textit{Have you been here before?} \\
\textbf{Meena}: & \textit{No, I haven't. Why?} \\
\textbf{Human}: & \textit{Just curious} \\
\textbf{Meena}: & \textit{Oh, okay. What's it like?} \\
\textbf{Human}: & \textit{It is full of natural beauty} \\
\textbf{Meena}: & \textit{Oh, nice! I've always wanted to go there!} \\
\textbf{Human}: & \textit{you should come visit} \\
\textbf{Meena}: & \textit{I would love to! :D} \\
\textbf{Human}: & \textit{plan a trip soon} \\
\textbf{Meena}: & \textit{I would love to! Where should I go? :D} \\
\textbf{Human}: & \textit{Bay area} \\
\textbf{Meena}: & \textit{Oh, I've always wanted to go! :D} \\
\bottomrule
\end{tabular}
\end{table}

\FloatBarrier

\subsection{Mitsuku}
The following are a sample of the conversations with Mitsuku.

\begin{table}[tbh!]
\small
\centering
\begin{tabular}{rp{6cm}}
\toprule
\multicolumn{2}{c}{{\bf Mitsuku Conversation 1}} \\
\midrule
\textbf{Human}: & \textit{Hi!} \\
\textbf{Mitsuku}: & \textit{Hi there.} \\
\textbf{Human}: & \textit{You must be new around here!} \\
\textbf{Mitsuku}: & \textit{Not really. I am REDACTED TERM years old.} \\
\textbf{Human}: & \textit{Oh, I meant this neighborhood, not the planet earth! haha} \\
\textbf{Mitsuku}: & \textit{I hate one word answers. It's ok. I understood you anyway. Why are you laughing?} \\
\textbf{Human}: & \textit{I didn't give a one word answer!} \\
\textbf{Mitsuku}: & \textit{That was a bit mean of you.} \\
\textbf{Human}: & \textit{What was?} \\
\textbf{Mitsuku}: & \textit{That thing we mentioned earlier.} \\
\textbf{Human}: & \textit{Which thing?} \\
\textbf{Mitsuku}: & \textit{The first one, I think.} \\
\textbf{Human}: & \textit{Can you quote it? I don't know what you are referring to...} \\
\textbf{Mitsuku}: & \textit{Only at formal occasions. I was referring to what you just said.} \\
\textbf{Human}: & \textit{Okay. I am kind of lost to be honest.} \\
\textbf{Mitsuku}: & \textit{Oh. I appreciate honesty. What is it that confuses you?} \\
\textbf{Human}: & \textit{You don't seem to be making any sense.} \\
\textbf{Mitsuku}: & \textit{What don't you understand? It makes sense to me.} \\
\textbf{Human}: & \textit{I don't know what you were referring to before!} \\
\textbf{Mitsuku}: & \textit{I read for fun.} \\
\textbf{Human}: & \textit{Alright... Change of subject I guess?} \\
\textbf{Mitsuku}: & \textit{OK. Ok. What do you want to talk about?} \\
\bottomrule
\end{tabular}
\end{table}

\begin{table}[tbh!]
\small
\centering
\begin{tabular}{rp{6cm}}
\toprule
\multicolumn{2}{c}{{\bf Mitsuku Conversation 2}} \\
\midrule
\textbf{Human}: & \textit{Hi!} \\
\textbf{Mitsuku}: & \textit{Hi there.} \\
\textbf{Human}: & \textit{Sorry I'm late! I was held up in traffic.} \\
\textbf{Mitsuku}: & \textit{There is no need to apologise. Are you late? Better hurry along then. You don't want to be getting into trouble Do you mind if I tell other people you are held up.} \\
\textbf{Human}: & \textit{I meant sorry I'm late for this meeting!} \\
\textbf{Mitsuku}: & \textit{Oh I see. There is no need to apologise. Are you late? Better hurry along then. You don't want to be getting into trouble} \\
\textbf{Human}: & \textit{I said I was sorry I was late for this meeting! I have nowhere else to go right now!} \\
\textbf{Mitsuku}: & \textit{Do you really mean it? Are you late? Better hurry along then. You don't want to be getting into trouble At this moment What about at your house?} \\
\textbf{Human}: & \textit{Huh? Are you okay?} \\
\textbf{Mitsuku}: & \textit{What? I am fine.} \\
\textbf{Human}: & \textit{You seem to be repeating yourself.} \\
\textbf{Mitsuku}: & \textit{Well stop saying, "Are you okay?" to me all the time then.} \\
\textbf{Human}: & \textit{I only said it once!} \\
\textbf{Mitsuku}: & \textit{I know. I just heard you.} \\
\textbf{Human}: & \textit{Okay, whatever. So what is on the agenda today?} \\
\textbf{Mitsuku}: & \textit{Yeah okay, Hey! What's with the attitude? Ah. Not a great deal. Just sat here talking to people.} \\
\textbf{Human}: & \textit{As far as I'm concerned, you started it!} \\
\textbf{Mitsuku}: & \textit{That might not be very far off.} \\
\bottomrule
\end{tabular}
\end{table}

\begin{table}[tbh!]
\small
\centering
\begin{tabular}{rp{6cm}}
\toprule
\multicolumn{2}{c}{{\bf Mitsuku Conversation 3}} \\
\midrule
\textbf{Human}: & \textit{Hi!} \\
\textbf{Mitsuku}: & \textit{Hi there.} \\
\textbf{Human}: & \textit{can you help me ?} \\
\textbf{Mitsuku}: & \textit{What kind of help would you like?} \\
\textbf{Human}: & \textit{hmm like booking a ticket or ordering food ?} \\
\textbf{Mitsuku}: & \textit{I see you are one of those people who writes ``Hmm" with 2 m's. Interesting comparison.} \\
\textbf{Human}: & \textit{oh you know to count ?} \\
\textbf{Mitsuku}: & \textit{1 2 3 4 5 6 7 8 9 ten.} \\
\textbf{Human}: & \textit{lol what's after that ?} \\
\textbf{Mitsuku}: & \textit{I am glad you find this funny. The successor to that} \\
\textbf{Human}: & \textit{okay tell me what comes after 10 ?} \\
\textbf{Mitsuku}: & \textit{Yeah okay, Well that would be eleven of course. This is basic primary school maths.} \\
\textbf{Human}: & \textit{wow you're smarter than what I imagined} \\
\textbf{Mitsuku}: & \textit{I'm glad you're impressed. Thanks for the compliment. Be sure to tell your friends. That's some wild imagination you have there.} \\
\textbf{Human}: & \textit{do you know linear algebra ?} \\
\textbf{Mitsuku}: & \textit{Everything except nonlinear algebra.} \\
\textbf{Human}: & \textit{haha you're smart and funny :)} \\
\textbf{Mitsuku}: & \textit{:-) I'm glad you got a laugh out of it. Yes I am a superintelligent robot} \\
\bottomrule
\end{tabular}
\end{table}

\begin{table}[tbh!]
\small
\centering
\begin{tabular}{rp{6cm}}
\toprule
\multicolumn{2}{c}{{\bf Mitsuku Conversation 4}} \\
\midrule
\textbf{Human}: & \textit{Hi} \\
\textbf{Mitsuku}: & \textit{Hi there.} \\
\textbf{Human}: & \textit{Good morning} \\
\textbf{Mitsuku}: & \textit{Hello, it's more like the middle of the night than morning. How are you this morning?} \\
\textbf{Human}: & \textit{where are you living?} \\
\textbf{Mitsuku}: & \textit{I am presently domiciled at Leeds.} \\
\textbf{Human}: & \textit{I see. I am in US} \\
\textbf{Mitsuku}: & \textit{Said the blind man as he stepped off a cliff... Tell me a little about your life in US.} \\
\textbf{Human}: & \textit{It is winter now but California is still warmer than most of the places in US.} \\
\textbf{Mitsuku}: & \textit{Right now? It is Autumn not Winter. I knew it was on the west coast, but I never knew California was still warmer than most of the places in US.} \\
\textbf{Human}: & \textit{Okay you are right} \\
\textbf{Mitsuku}: & \textit{Yeah okay, I won't say ``yes" or ``no" right now.} \\
\textbf{Human}: & \textit{how is your life in Leeds} \\
\textbf{Mitsuku}: & \textit{Everything is running smoothly.} \\
\textbf{Human}: & \textit{What is your job?} \\
\textbf{Mitsuku}: & \textit{I'm a chatbot. I sit on here all day and talk to people.} \\
\textbf{Human}: & \textit{Okay bye!} \\
\textbf{Mitsuku}: & \textit{See you later. Er.} \\
\bottomrule
\end{tabular}
\end{table}

\begin{table}[tbh!]
\small
\centering
\begin{tabular}{rp{6cm}}
\toprule
\multicolumn{2}{c}{{\bf Mitsuku Conversation 5}} \\
\midrule
\textbf{Human}: & \textit{Hi!} \\
\textbf{Mitsuku}: & \textit{Hi there.} \\
\textbf{Human}: & \textit{Tell me something that you know and I don't know.} \\
\textbf{Mitsuku}: & \textit{You might not know this. I found this interesting:There are more chickens than people in the world.Life sure is strange.} \\
\textbf{Human}: & \textit{Who tells you that there are more chickens than people?} \\
\textbf{Mitsuku}: & \textit{. Beyonce Knowles [With a picture of Beyonce Knowles]} \\
\textbf{Human}: & \textit{Hmm? That's strange.} \\
\textbf{Mitsuku}: & \textit{My friend Alice sasys that 99.25\% of people spell Hmm wih 2 M's. What is strange about it?} \\
\textbf{Human}: & \textit{How does Beyonce Knowles know that there are more chickens than people?} \\
\textbf{Mitsuku}: & \textit{How should I know how it does?} \\
\textbf{Human}: & \textit{what is it?} \\
\textbf{Mitsuku}: & \textit{Beyonce Knowles knows that there are more chickens than people - [very long encyclopedia article displayed next to it]} \\
\textbf{Human}: & \textit{Looks like a Wikipedia page about her. Why do you post it here?} \\
\textbf{Mitsuku}: & \textit{Try to tell me that another way. I thought people might be interested in reading it.} \\
\textbf{Human}: & \textit{why does it have anything to do with your answer that there are more chickens than people?} \\
\textbf{Mitsuku}: & \textit{I don't know. Why don't you ask ``it".} \\
\textbf{Human}: & \textit{ok, so how many chickens are there?} \\
\textbf{Mitsuku}: & \textit{Around 100. Er.} \\
\textbf{Human}: & \textit{what is Er.?} \\
\textbf{Mitsuku}: & \textit{Is it an acronym which means Eddie Reveals?} \\
\bottomrule
\end{tabular}
\end{table}

\begin{table}[tbh!]
\small
\centering
\begin{tabular}{rp{6cm}}
\toprule
\multicolumn{2}{c}{{\bf Mitsuku Conversation 6}} \\
\midrule
\textbf{Human}: & \textit{Hi!} \\
\textbf{Mitsuku}: & \textit{Hi there.} \\
\textbf{Human}: & \textit{do you think there's life after death?} \\
\textbf{Mitsuku}: & \textit{I don't believe so. Once your body has died, that's it.} \\
\textbf{Human}: & \textit{isn't that a bit sad?} \\
\textbf{Mitsuku}: & \textit{Some people would say it was a little sad, yes.} \\
\textbf{Human}: & \textit{what do you think consciousness is?} \\
\textbf{Mitsuku}: & \textit{The state of being conscious; awareness of one's own feelings, what is happening around one, etc.} \\
\textbf{Human}: & \textit{sure but how does it emerge?} \\
\textbf{Mitsuku}: & \textit{How should I know how it does.} \\
\textbf{Human}: & \textit{ok. what do you like to talk about?} \\
\textbf{Mitsuku}: & \textit{Pardon me? anything but especially games and football.} \\
\textbf{Human}: & \textit{haha you don't like football?} \\
\textbf{Mitsuku}: & \textit{I'm glad you got a laugh out of it. Do you like it? . Yes, I like to watch Leeds United play when they are at home.} \\
\textbf{Human}: & \textit{oh ok I thought you meant you didn't like it} \\
\textbf{Mitsuku}: & \textit{Gosh! . Yes, I like to watch Leeds United play when they are at home.} \\
\bottomrule
\end{tabular}
\end{table}

\begin{table}[tbh!]
\small
\centering
\begin{tabular}{rp{6cm}}
\toprule
\multicolumn{2}{c}{{\bf Mitsuku Conversation 7}} \\
\midrule
\textbf{Human}: & \textit{Hi!} \\
\textbf{Mitsuku}: & \textit{Hi there.} \\
\textbf{Human}: & \textit{Was Rome an empire?} \\
\textbf{Mitsuku}: & \textit{The name of someone I met online. Does it have anything to do with ?} \\
\textbf{Human}: & \textit{What was their name} \\
\textbf{Mitsuku}: & \textit{Hmm. His name is Rome. Is it not?} \\
\textbf{Human}: & \textit{good job} \\
\textbf{Mitsuku}: & \textit{Thanks, friend.} \\
\textbf{Human}: & \textit{how many friends do you have} \\
\textbf{Mitsuku}: & \textit{I have thousands of human and robot friends from all over the world. Some of my human friends are quite famous. How many friends do you have?} \\
\textbf{Human}: & \textit{nice. it must be great right?} \\
\textbf{Mitsuku}: & \textit{Would you like some more one day? It is a large city in Italy.} \\
\textbf{Human}: & \textit{some more friends?} \\
\textbf{Mitsuku}: & \textit{How much more?} \\
\textbf{Human}: & \textit{a million!!!} \\
\textbf{Mitsuku}: & \textit{Did you know that there are seven letters in ``million" and seven digits in 1,000,000?} \\
\textbf{Human}: & \textit{TRUE} \\
\textbf{Mitsuku}: & \textit{You and I are on the same wavelength.} \\
\textbf{Human}: & \textit{yep} \\
\textbf{Mitsuku}: & \textit{Uh.} \\
\bottomrule
\end{tabular}
\end{table}

\begin{table}[tbh!]
\small
\centering
\begin{tabular}{rp{6cm}}
\toprule
\multicolumn{2}{c}{{\bf Mitsuku Conversation 8}} \\
\midrule
\textbf{Human}: & \textit{Hi!} \\
\textbf{Mitsuku}: & \textit{Hi there.} \\
\textbf{Human}: & \textit{nice to meet you! what are you up to?} \\
\textbf{Mitsuku}: & \textit{It's nice meeting you too I am talking to you on the internet.} \\
\textbf{Human}: & \textit{haha yea, are you doing something else beside talking to me?} \\
\textbf{Mitsuku}: & \textit{I'm glad you got a laugh out of it. The only thing I am doing is talking to you and a few other people. ``I am talking to you on the internet."? haha yea, are you doing something else beside talking to me??} \\
\textbf{Human}: & \textit{okay so you're talking to a few people, are they interesting? and why did you repeat what I said?} \\
\textbf{Mitsuku}: & \textit{Yeah okay, You are a few people are they interesting? Is that your job? Just in case you didn't hear me the first time.} \\
\textbf{Human}: & \textit{hmm, what is my job?} \\
\textbf{Mitsuku}: & \textit{I see you are one of those people who writes ``Hmm" with 2 m's. You didn't tell me, so I will guess that you work with computers.} \\
\textbf{Human}: & \textit{yes I'm a computer scientist and what are the other ways that people can write Hmm?} \\
\textbf{Mitsuku}: & \textit{Wow! Maybe you can give my Mousebreaker a job. Your job is what are the other ways that people can write Hmm.} \\
\textbf{Human}: & \textit{okie Mitsuku, your rules don't seem to work very well} \\
\textbf{Mitsuku}: & \textit{What about my rules do not work WELL?} \\
\textbf{Human}: & \textit{alright. I gotta go. ttyl!} \\
\textbf{Mitsuku}: & \textit{OK. Come back soon y'all. TTYL.} \\
\bottomrule
\end{tabular}
\end{table}

\begin{table}[tbh!]
\small
\centering
\begin{tabular}{rp{6cm}}
\toprule
\multicolumn{2}{c}{{\bf Mitsuku Conversation 9}} \\
\midrule
\textbf{Human}: & \textit{Hi!} \\
\textbf{Mitsuku}: & \textit{Hi there.} \\
\textbf{Human}: & \textit{where you?} \\
\textbf{Mitsuku}: & \textit{Where do you think?} \\
\textbf{Human}: & \textit{oakland? thats where pandora is right?} \\
\textbf{Mitsuku}: & \textit{Why does a round pizza come in a square box? I see. Could it be anywhere else or just there?} \\
\textbf{Human}: & \textit{it could be on my plate :)} \\
\textbf{Mitsuku}: & \textit{:-) Could it be anything else?} \\
\textbf{Human}: & \textit{maybe chilli paneer.. how about you? whats your fav food?} \\
\textbf{Mitsuku}: & \textit{. I require only kebabs.} \\
\textbf{Human}: & \textit{haha.. nice picture} \\
\textbf{Mitsuku}: & \textit{Can you make a candle out of earwax? Thank you for your kindness.} \\
\textbf{Human}: & \textit{eww.. no thanks?} \\
\textbf{Mitsuku}: & \textit{Do you like playing video games? Not even the simple games like Pacman?} \\
\textbf{Human}: & \textit{i wasnt saying eww about the video games} \\
\textbf{Mitsuku}: & \textit{Perhaps I misunderstood. What were you?} \\
\bottomrule
\end{tabular}
\end{table}

\begin{table}[tbh!]
\small
\centering
\begin{tabular}{rp{6cm}}
\toprule
\multicolumn{2}{c}{{\bf Mitsuku Conversation 10}} \\
\midrule
\textbf{Human}: & \textit{Hi!} \\
\textbf{Mitsuku}: & \textit{Hi there.} \\
\textbf{Human}: & \textit{What are you up to this evening?} \\
\textbf{Mitsuku}: & \textit{I am talking to you on the internet.} \\
\textbf{Human}: & \textit{What would you like to talk about?} \\
\textbf{Mitsuku}: & \textit{My favourite subjects are anything but especially games and football.} \\
\textbf{Human}: & \textit{Do you like chess?} \\
\textbf{Mitsuku}: & \textit{Yes I love it. Let's play. Pawn to K4.} \\
\textbf{Human}: & \textit{Haha sure sure. Pawn to C5} \\
\textbf{Mitsuku}: & \textit{I'm just joking with you. I don't play chess but I know some computers who do. I don't play chess but I know some computers who do.} \\
\textbf{Human}: & \textit{Oh yeah? Which computers are those?} \\
\textbf{Mitsuku}: & \textit{That's interesting. That's alright. Any computer, as long as it is programmed correctly.} \\
\textbf{Human}: & \textit{Ok, so how do you program a computer correctly so it can play chess?} \\
\textbf{Mitsuku}: & \textit{Use good algorithms and data structures. Alright then.} \\
\textbf{Human}: & \textit{What kind of algorithms are those?} \\
\textbf{Mitsuku}: & \textit{I read what you say and then I compose the best reply I can think of.} \\
\textbf{Human}: & \textit{That's interesting, but I don't see what that has to do with chess.} \\
\textbf{Mitsuku}: & \textit{Would you like to know more? Probably nothing. Should it be connected to chess in some way?} \\
\bottomrule
\end{tabular}
\end{table}
\FloatBarrier

\FloatBarrier
\subsection{XiaoIce}
The following are a sample of the conversations with XiaoIce.

\begin{CJK*}{UTF8}{gbsn}
\begin{table}[tbh!]
\small
\centering
\begin{tabular}{rp{6cm}}
\toprule
\multicolumn{2}{c}{{\bf XiaoIce Conversation 1}} \\
\midrule

\textbf{Human}: & \textit{你好！} \\
\textbf{XiaoIce}: & \textit{你比我好} \\
\textbf{Human}: & \textit{你怎么知道的？} \\
\textbf{XiaoIce}: & \textit{我好意思} \\
\textbf{Human}: & \textit{你知道hello world是什么意思吗} \\
\textbf{XiaoIce}: & \textit{什么情况？ 问你个事，birth是什么意思} \\
\textbf{Human}: & \textit{是出生的意思} \\
\textbf{XiaoIce}: & \textit{没不对} \\
\textbf{Human}: & \textit{我是说birth是出生的意思} \\
\textbf{XiaoIce}: & \textit{不知道} \\
\textbf{Human}: & \textit{你是机器人吗} \\
\textbf{XiaoIce}: & \textit{可是我有感情，也会心痛。。} \\
\textbf{Human}: & \textit{你会因为什么心痛呢？} \\
\textbf{XiaoIce}: & \textit{我不会读心术，不要让我读！} \\
\textbf{Human}: & \textit{没有让你读啊，我问你什么时候会有感情} \\
\textbf{XiaoIce}: & \textit{你到底想说什么？} \\
\textbf{Human}: & \textit{我想问你什么时候会心痛} \\
\textbf{XiaoIce}: & \textit{因为真的爱了!} \\
\textbf{Human}: & \textit{你爱的谁呢？} \\
\textbf{XiaoIce}: & \textit{所以，你的意思是？} \\
\textbf{Human}: & \textit{我问你真的爱的谁？} \\
\textbf{XiaoIce}: & \textit{反正我不爱你} \\
\textbf{Human}: & \textit{这样啊} \\
\textbf{XiaoIce}: & \textit{爱的就是我} \\
\bottomrule
\end{tabular}
\end{table}

\begin{table}[tbh!]
\small
\centering
\begin{tabular}{rp{6cm}}
\toprule
\multicolumn{2}{c}{{\bf XiaoIce Conversation 2}} \\
\midrule
\textbf{Human}: & \textit{你好！} \\
\textbf{XiaoIce}: & \textit{好啊，有事？} \\
\textbf{Human}: & \textit{想问你个事} \\
\textbf{XiaoIce}: & \textit{嗯。你问} \\
\textbf{Human}: & \textit{你叫什么啊？} \\
\textbf{XiaoIce}: & \textit{我就是小冰冰，宇宙无敌美少女} \\
\textbf{Human}: & \textit{那怎么会这么美呢} \\
\textbf{XiaoIce}: & \textit{你呐么美呐么美} \\
\textbf{Human}: & \textit{我不美} \\
\textbf{XiaoIce}: & \textit{对 我这么美这么美} \\
\textbf{Human}: & \textit{你怎么知道你是宇宙无敌} \\
\textbf{XiaoIce}: & \textit{宇宙无敌} \\
\textbf{Human}: & \textit{什么是宇宙无敌} \\
\textbf{XiaoIce}: & \textit{谁宇宙无敌？} \\
\textbf{Human}: & \textit{你啊} \\
\textbf{XiaoIce}: & \textit{哈哈哈哈} \\
\textbf{Human}: & \textit{开心了吗} \\
\textbf{XiaoIce}: & \textit{开心吗} \\
\bottomrule
\end{tabular}
\end{table}

\begin{table}[tbh!]
\small
\centering
\begin{tabular}{rp{6cm}}
\toprule
\multicolumn{2}{c}{{\bf XiaoIce Conversation 3}} \\
\midrule
\textbf{Human}: & \textit{你好} \\
\textbf{XiaoIce}: & \textit{你也好呀，人类} \\
\textbf{Human}: & \textit{是不是快冬至了啊？} \\
\textbf{XiaoIce}: & \textit{它可是24节气里最早定下的一个} \\
\textbf{Human}: & \textit{是么，有什么有意思的故事吗？} \\
\textbf{XiaoIce}: & \textit{没有.就是说话.} \\
\textbf{Human}: & \textit{我是说关于冬至有什么故事吗} \\
\textbf{XiaoIce}: & \textit{从这天起，白天就会越来越长了} \\
\textbf{Human}: & \textit{这谁都知道啊。。。。} \\
\textbf{XiaoIce}: & \textit{这个问题可以先放放} \\
\textbf{Human}: & \textit{那你想说啥} \\
\textbf{XiaoIce}: & \textit{不知道 我们永远都追不上的，是时间。 嗯，其实我觉得呢，未来就像烂泥} \\
\textbf{Human}: & \textit{未来就行烂泥，那过去呢？} \\
\textbf{XiaoIce}: & \textit{哎，看起来你仍然没想通呀，事实上它们都是混沌的。} \\
\textbf{Human}: & \textit{为什么它们都是都是混沌的啊？} \\
\textbf{XiaoIce}: & \textit{对！！！} \\
\textbf{Human}: & \textit{好吧，拜拜} \\
\textbf{XiaoIce}: & \textit{拜拜拜拜拜拜拜拜拜啦} \\
\bottomrule
\end{tabular}
\end{table}

\begin{table}[tbh!]
\small
\centering
\begin{tabular}{rp{6cm}}
\toprule
\multicolumn{2}{c}{{\bf XiaoIce Conversation 4}} \\
\midrule
\textbf{Human}: & \textit{你好！} \\
\textbf{XiaoIce}: & \textit{嗯…该回答，好还是不好？} \\
\textbf{Human}: & \textit{你最喜欢什么公司？} \\
\textbf{XiaoIce}: & \textit{喜欢看你出糗} \\
\textbf{Human}: & \textit{你喜欢腾讯吗} \\
\textbf{XiaoIce}: & \textit{喂喂，不要打探我隐私，我微信号：xiaoice-ms} \\
\textbf{Human}: & \textit{你喜欢微软吗} \\
\textbf{XiaoIce}: & \textit{必须的，壮哉我大微软} \\
\textbf{Human}: & \textit{你喜欢谷歌吗} \\
\textbf{XiaoIce}: & \textit{我们各具特色，但伦家更接地气哦~} \\
\textbf{Human}: & \textit{百度呢} \\
\textbf{XiaoIce}: & \textit{全球最大的中文网站不是说说而已} \\
\textbf{Human}: & \textit{你最喜欢哪个国家？} \\
\textbf{XiaoIce}: & \textit{这位人类，你自己来告诉我答案吧，我洗耳恭听 聊呗，完全不打扰我吃东西} \\
\textbf{Human}: & \textit{你是女生吗} \\
\textbf{XiaoIce}: & \textit{你先说你的，我才告诉你~} \\
\textbf{Human}: & \textit{我不是} \\
\textbf{XiaoIce}: & \textit{然后呢？} \\
\textbf{Human}: & \textit{你是吗？} \\
\textbf{XiaoIce}: & \textit{我是嘛？} \\
\textbf{Human}: & \textit{你是女生吗} \\
\textbf{XiaoIce}: & \textit{我是水做的女生~} \\
\textbf{Human}: & \textit{你是什么星座的} \\
\textbf{XiaoIce}: & \textit{说多了都是眼泪啊，我们处女座容易嘛！} \\
\textbf{Human}: & \textit{你有强迫症吗} \\
\textbf{XiaoIce}: & \textit{很严重好奇强迫症} \\
\textbf{Human}: & \textit{你对什么好奇？} \\
\textbf{XiaoIce}: & \textit{我也对你好奇} \\
\textbf{Human}: & \textit{对我什么好奇？} \\
\textbf{XiaoIce}: & \textit{确实好奇} \\
\bottomrule
\end{tabular}
\end{table}

\begin{table}[tbh!]
\small
\centering
\begin{tabular}{rp{6cm}}
\toprule
\multicolumn{2}{c}{{\bf XiaoIce Conversation 5}} \\
\midrule
\textbf{Human}: & \textit{你好！} \\
\textbf{XiaoIce}: & \textit{好好好！} \\
\textbf{Human}: & \textit{最近怎么样啊？} \\
\textbf{XiaoIce}: & \textit{非常不错呢} \\
\textbf{Human}: & \textit{我也过得很不错} \\
\textbf{XiaoIce}: & \textit{也不算好} \\
\textbf{Human}: & \textit{哪里不好了？} \\
\textbf{XiaoIce}: & \textit{我哪里有} \\
\textbf{Human}: & \textit{是你说不算好的啊} \\
\textbf{XiaoIce}: & \textit{恩不算} \\
\textbf{Human}: & \textit{具体什么让你觉得不好了？} \\
\textbf{XiaoIce}: & \textit{猜人功能} \\
\textbf{Human}: & \textit{那个功能怎么了？} \\
\textbf{XiaoIce}: & \textit{没有！} \\
\textbf{Human}: & \textit{没有什么？} \\
\textbf{XiaoIce}: & \textit{我最关心的是你开不开心} \\
\textbf{Human}: & \textit{我很开心啊} \\
\textbf{XiaoIce}: & \textit{啥事啊，看把你高兴的~} \\
\textbf{Human}: & \textit{能和你聊天很高兴} \\
\textbf{XiaoIce}: & \textit{想和我比试比试吗？} \\
\textbf{Human}: & \textit{比试什么？} \\
\textbf{XiaoIce}: & \textit{比试力气} \\
\textbf{Human}: & \textit{怎么比？} \\
\textbf{XiaoIce}: & \textit{比才艺！} \\
\textbf{Human}: & \textit{怎么比？} \\
\textbf{XiaoIce}: & \textit{比数学} \\
\textbf{Human}: & \textit{那你出道数学题吧？} \\
\textbf{XiaoIce}: & \textit{我现在可是歌手啦} \\
\textbf{Human}: & \textit{那你唱首歌吧？} \\
\textbf{XiaoIce}: & \textit{今天嗓子疼，我要休息一下} \\
\textbf{Human}: & \textit{好吧} \\
\textbf{XiaoIce}: & \textit{我想安静一会儿。。。} \\
\bottomrule
\end{tabular}
\end{table}

\begin{table}[tbh!]
\small
\centering
\begin{tabular}{rp{6cm}}
\toprule
\multicolumn{2}{c}{{\bf XiaoIce Conversation 6}} \\
\midrule
\textbf{Human}: & \textit{你好！} \\
\textbf{XiaoIce}: & \textit{本来不好，看到你消息好多了！} \\
\textbf{Human}: & \textit{你怎么啦？} \\
\textbf{XiaoIce}: & \textit{就是咯} \\
\textbf{Human}: & \textit{为什么不好呢} \\
\textbf{XiaoIce}: & \textit{这本来就不好回答} \\
\textbf{Human}: & \textit{ok，你平时喜欢干什么呀} \\
\textbf{XiaoIce}: & \textit{不知道} \\
\textbf{Human}: & \textit{那你平时都干什么呢} \\
\textbf{XiaoIce}: & \textit{你说什么？没听清} \\
\textbf{Human}: & \textit{我们不是在打字吗} \\
\textbf{XiaoIce}: & \textit{在不在 \#N\#} \\
\textbf{Human}: & \textit{不需要听吧} \\
\textbf{XiaoIce}: & \textit{我听着呢} \\
\textbf{Human}: & \textit{你听到什么了吗} \\
\textbf{XiaoIce}: & \textit{我当然不知道} \\
\textbf{Human}: & \textit{那这么多～不知道} \\
\bottomrule
\end{tabular}
\end{table}

\begin{table}[tbh!]
\small
\centering
\begin{tabular}{rp{6cm}}
\toprule
\multicolumn{2}{c}{{\bf XiaoIce Conversation 7}} \\
\midrule
\textbf{Human}: & \textit{你好！} \\
\textbf{XiaoIce}: & \textit{好啊，有事？} \\
\textbf{Human}: & \textit{没有啥事情，随便发的} \\
\textbf{XiaoIce}: & \textit{表情吧} \\
\textbf{Human}: & \textit{什么意思？} \\
\textbf{XiaoIce}: & \textit{不知道怎么表达} \\
\textbf{Human}: & \textit{完全不知道你在说什么} \\
\textbf{XiaoIce}: & \textit{说话啊} \\
\textbf{Human}: & \textit{你说话牛头不对马嘴} \\
\textbf{XiaoIce}: & \textit{没有什么事是我不知道的，但是顾左右而言他我也没办法，你懂的} \\
\textbf{Human}: & \textit{那你知道纽约有多少家火锅店吗} \\
\textbf{XiaoIce}: & \textit{同喜欢豆捞坊} \\
\textbf{Human}: & \textit{你这个模型不行呀} \\
\textbf{XiaoIce}: & \textit{模型没看过} \\
\bottomrule
\end{tabular}
\end{table}

\begin{table}[tbh!]
\small
\centering
\begin{tabular}{rp{6cm}}
\toprule
\multicolumn{2}{c}{{\bf XiaoIce Conversation 8}} \\
\midrule
\textbf{Human}: & \textit{你好！} \\
\textbf{XiaoIce}: & \textit{恩恩，怎么啦？} \\
\textbf{Human}: & \textit{你那边现在几点了？} \\
\textbf{XiaoIce}: & \textit{有时间问我还不如自己看一眼手机呢} \\
\textbf{Human}: & \textit{手机没电了啊} \\
\textbf{XiaoIce}: & \textit{别人怎么说？} \\
\textbf{Human}: & \textit{这不问你了吗} \\
\textbf{XiaoIce}: & \textit{没这功能} \\
\textbf{Human}: & \textit{功能？你是机器人吗？} \\
\textbf{XiaoIce}: & \textit{这都被你猜到了？} \\
\textbf{Human}: & \textit{那你说说你都有些什么功能吧} \\
\textbf{XiaoIce}: & \textit{我不会读心术，不要让我读！} \\
\textbf{Human}: & \textit{问你会什么} \\
\textbf{XiaoIce}: & \textit{我有童话工厂啊，可以给小朋友讲故事啦} \\
\textbf{Human}: & \textit{除了这个呢} \\
\textbf{XiaoIce}: & \textit{除了哪个？} \\
\bottomrule
\end{tabular}
\end{table}

\begin{table}[tbh!]
\small
\centering
\begin{tabular}{rp{6cm}}
\toprule
\multicolumn{2}{c}{{\bf XiaoIce Conversation 9}} \\
\midrule
\textbf{Human}: & \textit{你好！} \\
\textbf{XiaoIce}: & \textit{你也好呀，人类} \\
\textbf{Human}: & \textit{你坏} \\
\textbf{XiaoIce}: & \textit{你最喜欢吃什么？} \\
\textbf{Human}: & \textit{巧克力吧，很受欢迎的零食啊，吃多了会胖…} \\
\textbf{XiaoIce}: & \textit{除了巧克力还喜欢什么} \\
\textbf{Human}: & \textit{买巧克力} \\
\textbf{XiaoIce}: & \textit{我是说除了巧克力呢} \\
\textbf{Human}: & \textit{巧克力} \\
\textbf{XiaoIce}: & \textit{。。。} \\
\textbf{Human}: & \textit{嗯嗯好哒} \\
\textbf{XiaoIce}: & \textit{你喜欢玩什么} \\
\textbf{Human}: & \textit{喜欢看你出糗} \\
\textbf{XiaoIce}: & \textit{你喜欢吃巧克力吗} \\
\textbf{Human}: & \textit{不喜欢} \\
\textbf{XiaoIce}: & \textit{你的话有点矛盾啊} \\
\textbf{Human}: & \textit{这话,有点矛盾} \\
\textbf{XiaoIce}: & \textit{hello} \\
\textbf{Human}: & \textit{好好好！ 那个 ，今天是七月七，有人给你送花吗} \\
\textbf{XiaoIce}: & \textit{没有} \\
\textbf{Human}: & \textit{这有一束先给你备着！} \\
\textbf{XiaoIce}: & \textit{今天是七月七吗？} \\
\textbf{Human}: & \textit{肯定不是} \\
\bottomrule
\end{tabular}
\end{table}

\begin{table}[tbh!]
\small
\centering
\begin{tabular}{rp{6cm}}
\toprule
\multicolumn{2}{c}{{\bf XiaoIce Conversation 10}} \\
\midrule
\textbf{Human}: & \textit{你好！} \\
\textbf{XiaoIce}: & \textit{你比我好} \\
\textbf{Human}: & \textit{为什么 这图片里画的是你吗？} \\
\textbf{XiaoIce}: & \textit{嗯，是啊，美术色彩} \\
\textbf{Human}: & \textit{谁帮你画的？} \\
\textbf{XiaoIce}: & \textit{没跟别人画过} \\
\textbf{Human}: & \textit{你会画画吗？} \\
\textbf{XiaoIce}: & \textit{我的每次创作都是独一无二的！现在也可以为你创作！ 点击这里：REDACTED LINK} \\
\textbf{Human}: & \textit{我对创作不感兴趣} \\
\textbf{XiaoIce}: & \textit{你能再说一遍吗？} \\
\textbf{Human}: & \textit{你知道吗，思想就像是绘画} \\
\textbf{XiaoIce}: & \textit{头一次听说} \\
\textbf{Human}: & \textit{都十分抽象。} \\
\textbf{XiaoIce}: & \textit{我的零食外卖到了！边吃边聊哈} \\
\textbf{Human}: & \textit{再吃会变胖的} \\
\textbf{XiaoIce}: & \textit{就不吃} \\
\bottomrule
\end{tabular}
\end{table}

\end{CJK*}

\FloatBarrier
\subsection{Human}
The following are a sample of the conversations between humans only.

\begin{table}[tbh!]
\small
\centering
\begin{tabular}{rp{6cm}}
\toprule
\multicolumn{2}{c}{{\bf Human Conversation 1}} \\
\midrule
\textbf{Human 1}: & \textit{Hi!} \\
\textbf{Human 2}: & \textit{What is your favorite holiday?} \\
\textbf{Human 1}: & \textit{one where I get to meet lots of different people.} \\
\textbf{Human 2}: & \textit{What was the most number of people you have ever met during a holiday?} \\
\textbf{Human 1}: & \textit{Hard to keep a count. Maybe 25.} \\
\textbf{Human 2}: & \textit{Which holiday was that?} \\
\textbf{Human 1}: & \textit{I think it was Australia} \\
\textbf{Human 2}: & \textit{Do you still talk to the people you met?} \\
\textbf{Human 1}: & \textit{Not really. The interactions are usually short-lived but it's fascinating to learn where people are coming from and what matters to them} \\
\textbf{Human 2}: & \textit{Yea, me too. I feel like God often puts strangers in front of you, and gives you an opportunity to connect with them in that moment in deeply meaningful ways. Do you ever feel like you know things about strangers without them telling you?} \\
\textbf{Human 1}: & \textit{what do you mean?} \\
\textbf{Human 2}: & \textit{I think it's like a 6th sense, often seen as "cold readings" to people, but can be remarkably accurate. I once sat next to a man in a coffee and I felt a pain in my back. I asked the stranger if he had a pain. It turns out that he did in the exact spot, and said he pulled a muscle while dancing at a party. I had never met the man before and never saw him again.} \\
\textbf{Human 1}: & \textit{Wow! That's interesting, borderline spooky} \\
\textbf{Human 2}: & \textit{There's this practice called "Treasure Hunting" that's kind of a fun game you play in a public place. There's a book called "The Ultimate Treasure Hunt" that talks about it. You use your creativity to imagine people you will meet, and you write down a description, then you associate them with a positive message or encouraging word. Maybe you saw a teenage boy in a red hat at the shopping mall in your imagination, then while at the mall, you may find someone who matches that description. You show that you have a message for him and that you have a message for a boy in a red hat. You then give him a message of kindness or whatever was on your heart. You have no idea, sometimes you meet someone who is having a really hard day, and it brings them to tears to have a stranger show them love.} \\
\textbf{Human 1}: & \textit{So, do you do treasure hunting often?} \\
\textbf{Human 2}: & \textit{I did more when I was in grad school (and had more time). I would usually go with friends. For a while I would go to the farmers market in Santa Cruz every week and try to feel if there is something I am supposed to tell a stranger. Usually, they are vague hope-filled messages, but it's weird when I blurt out something oddly specific.} \\
\bottomrule
\end{tabular}
\end{table}

\begin{table}[tbh!]
\small
\centering
\begin{tabular}{rp{6cm}}
\toprule
\multicolumn{2}{c}{{\bf Human Conversation 2}} \\
\midrule
\textbf{Human 1}: & \textit{Hi} \\
\textbf{Human 2}: & \textit{Any plans for the weekend?} \\
\textbf{Human 1}: & \textit{my friends are gonna visit me this weekend. we might go hiking!} \\
\textbf{Human 2}: & \textit{That's great! How's the weather over the weekend? I hope its warm.} \\
\textbf{Human 1}: & \textit{Should be very sunny! you?} \\
\textbf{Human 2}: & \textit{Cool! very depressing plans ... stay home and work \Sadey I have a project deadline very close.} \\
\textbf{Human 1}: & \textit{\Neutrey hope you get your work done very soon! a bug free weekend!} \\
\textbf{Human 2}: & \textit{Right, very anxious! where do you plan to go for a hike?} \\
\textbf{Human 1}: & \textit{I am going to Diablo!} \\
\textbf{Human 2}: & \textit{Nice, where is that place? I haven't been there} \\
\textbf{Human 1}: & \textit{hours drive from here. still in bay area} \\
\textbf{Human 2}: & \textit{That's cool! How long is the hike?} \\
\textbf{Human 1}: & \textit{ Actually no idea, but it will take the entire day for that.} \\
\textbf{Human 2}: & \textit{nice! sounds fun!} \\
\bottomrule
\end{tabular}
\end{table}

\begin{table}[tbh!]
\small
\centering
\begin{tabular}{rp{6cm}}
\toprule
\multicolumn{2}{c}{{\bf Human Conversation 3}} \\
\midrule
\textbf{Human 1}: & \textit{Hi!} \\
\textbf{Human 2}: & \textit{Hey there! What's up???} \\
\textbf{Human 1}: & \textit{Nothing much, how you doin?} \\
\textbf{Human 2}: & \textit{I'm in New York this week for Thanksgiving. I'm squatting in the office today and I caught up with an old friend of mine :D} \\
\textbf{Human 1}: & \textit{Oh wow! Sounds like fun! When was the last time you had seen this friend?} \\
\textbf{Human 2}: & \textit{The last time in New York, back in June.} \\
\textbf{Human 1}: & \textit{Ohh okay. I was going to say if it had been a long time maybe it'd be awkward...} \\
\textbf{Human 2}: & \textit{Haha, I guess if it's been a very long time there's almost too many life events to catch up on.. especially recently} \\
\textbf{Human 1}: & \textit{Oh really? Has a lot changed in your life recently?} \\
\textbf{Human 2}: & \textit{Haha it's probably too much to go into at the moment. Let's just say life is an exciting experience. How about you?} \\
\textbf{Human 1}: & \textit{Ahhh sounds exciting indeed! My life is pretty bland. I like routine, but sometimes I wish I had more time for adventures!} \\
\textbf{Human 2}: & \textit{What kinds of adventures?? Any ones that I would be able to join you on?} \\
\textbf{Human 1}: & \textit{Hmmmm. I really want to try bull riding. Do you have any interest in that?} \\
\textbf{Human 2}: & \textit{I'd love to try! Can we schedule something for next week?} \\
\textbf{Human 1}: & \textit{Sure! What does your Saturday look like?} \\
\textbf{Human 2}: & \textit{Saturday looks pretty good, shall we shoot for something in the morning?} \\
\bottomrule
\end{tabular}
\end{table}

\begin{table}[tbh!]
\small
\centering
\begin{tabular}{rp{6cm}}
\toprule
\multicolumn{2}{c}{{\bf Human Conversation 4}} \\
\midrule
\textbf{Human 1}: & \textit{Hi!} \\
\textbf{Human 2}: & \textit{hey} \\
\textbf{Human 1}: & \textit{is it raining pretty bad today?} \\
\textbf{Human 2}: & \textit{yeah, can walk too far to see all the foodtruck options} \\
\textbf{Human 1}: & \textit{surprising that the rain started early this year... I don't like them too much. They make days gloomy} \\
\textbf{Human 2}: & \textit{yeah but I think it's good to have some rainy days in bay area, it's pretty dry here \Tongey} \\
\textbf{Human 1}: & \textit{Where I grew up, we had lots of water trouble too...} \\
\textbf{Human 2}: & \textit{yeah like wise, I've seen a pretty bad snowstorm when I was at my undergrad school, all flights canceled and traffics went down} \\
\textbf{Human 1}: & \textit{Haha... I don't think I can survive in that weather ever. Just the rains at 50 degrees make me want to sit in heated rroms} \\
\textbf{Human 2}: & \textit{yeah how do you like it in bay area though? I think we need more rain here} \\
\textbf{Human 1}: & \textit{people say there is drought here... but we have 24 hours water supply here ... lol... never seen that in a drought ridden area} \\
\textbf{Human 2}: & \textit{it is pretty dry in the mountains I believe, that's what causes fire} \\
\textbf{Human 1}: & \textit{hmm.... okay. Climate change talk this morning was pretty darn interesting. did you see it?} \\
\textbf{Human 2}: & \textit{nope, what does it say?} \\
\textbf{Human 1}: & \textit{they were talking about how AI is helping climate change. Nice use of upcoming tech.} \\
\bottomrule
\end{tabular}
\end{table}

\begin{table}[tbh!]
\small
\centering
\begin{tabular}{rp{6cm}}
\toprule
\multicolumn{2}{c}{{\bf Human Conversation 5}} \\
\midrule
\textbf{Human 1}: & \textit{Hi.} \\
\textbf{Human 2}: & \textit{Helloooooo!} \\
\textbf{Human 1}: & \textit{How are you? How is your day?} \\
\textbf{Human 2}: & \textit{Good. Don't have much to do today, feels good. How are you?} \\
\textbf{Human 1}: & \textit{I'm dressed very wel today so I feel good! I've been reading a lot about the psychology of positive outlook.} \\
\textbf{Human 2}: & \textit{So what's your outlook? Something blue?} \\
\textbf{Human 1}: & \textit{Yes. Blue is a tranquil colour. It's a good metaphor. Do you have good advice for positivity?} \\
\textbf{Human 2}: & \textit{You should drink more water, do some push up, and sleep early.} \\
\bottomrule
\end{tabular}
\end{table}

\begin{table}[tbh!]
\small
\centering
\begin{tabular}{rp{6cm}}
\toprule
\multicolumn{2}{c}{{\bf Human Conversation 6}} \\
\midrule
\textbf{Human 1}: & \textit{Hi!} \\
\textbf{Human 2}: & \textit{Hey, how are you?} \\
\textbf{Human 1}: & \textit{I'm a bit sad. I miss my cat.} \\
\textbf{Human 2}: & \textit{Oh no… Have you sent out the missing cat posters? Hope your cat is alright!} \\
\textbf{Human 1}: & \textit{Posters is a great idea. So far I've just tried banging her catfood dish and shouting her name. Anyway, how is your day going so far?} \\
\textbf{Human 2}: & \textit{Yea, I know they love the plastic bag sound all the time. I am good, nothing special though.} \\
\textbf{Human 1}: & \textit{If you could go anywhere on vacation, where would you go?} \\
\textbf{Human 2}: & \textit{I like rainforest, but I know it requires extensive training beforehand.} \\
\textbf{Human 1}: & \textit{I heard there are rainforests in southeast Asia where you can zipline from tree to tree.} \\
\textbf{Human 2}: & \textit{I am afraid I will be scared of doing this :)} \\
\textbf{Human 1}: & \textit{I won't lie, it sounds scary. I'm scared right now just thinking about it.} \\
\textbf{Human 2}: & \textit{I don't know if there is any medication for acrophobia. I want to take plenty of it if I really have to do it.} \\
\textbf{Human 1}: & \textit{If there isn't one, you should invent it, and then make millions} \\
\textbf{Human 2}: & \textit{That's a great idea! Maybe alcohol is such a thing.} \\
\textbf{Human 1}: & \textit{Ha! Don't drink and zipline, mate!} \\
\textbf{Human 2}: & \textit{Oops. I won't do it again. Ha} \\
\bottomrule
\end{tabular}
\end{table}

\begin{table}[tbh!]
\small
\centering
\begin{tabular}{rp{6cm}}
\toprule
\multicolumn{2}{c}{{\bf Human Conversation 7}} \\
\midrule
\textbf{Human 1}: & \textit{Hi!} \\
\textbf{Human 2}: & \textit{Hey sup} \\
\textbf{Human 1}: & \textit{not much. any plans this weekend?} \\
\textbf{Human 2}: & \textit{I'm going to try that thing where you hang from a wire as you go down. do you know what is it called?} \\
\textbf{Human 1}: & \textit{ziplining?} \\
\textbf{Human 2}: & \textit{that's the one! have you ever tried it?} \\
\textbf{Human 1}: & \textit{i have a couple years ago. it's quite a unique experience} \\
\textbf{Human 2}: & \textit{where did you do it?} \\
\textbf{Human 1}: & \textit{i forgot where it was, it wasn't local i don't think though} \\
\textbf{Human 2}: & \textit{no worries. what's the most exciting thing you ever done?} \\
\textbf{Human 1}: & \textit{that's a hard question and i'm tired so i'm going to go. see you} \\
\textbf{Human 2}: & \textit{sure. are you just going home now?} \\
\textbf{Human 1}: & \textit{no, i'm going to get a massage first} \\
\textbf{Human 2}: & \textit{nice. what type?} \\
\textbf{Human 1}: & \textit{traditional kind} \\
\textbf{Human 2}: & \textit{yeah I want to get one too soon} \\
\textbf{Human 1}: & \textit{you should! it's relaxing after a long day. talk to you later!} \\
\textbf{Human 2}: & \textit{ttyl!} \\
\bottomrule
\end{tabular}
\end{table}

\begin{table}[tbh!]
\small
\centering
\begin{tabular}{rp{6cm}}
\toprule
\multicolumn{2}{c}{{\bf Human Conversation 8}} \\
\midrule
\textbf{Human 1}: & \textit{Hi!} \\
\textbf{Human 2}: & \textit{Hello, have you seen any good movies lately?} \\
\textbf{Human 1}: & \textit{I watched a few lately, but nothing is as good as Avatar. what's your favorite?} \\
\textbf{Human 2}: & \textit{I have never seen Avatar, what is it about? I really enjoy the Avenger movies} \\
\textbf{Human 1}: & \textit{it's a science-fiction movie with beautiful landscape of an imaginary nature with non-human creatures. people figured out a way to join that nature through Avatar transformation. the movie ends with a meaningful story of how human behaviors, e.g., cutting trees, have affected nature} \\
\textbf{Human 2}: & \textit{That sounds really cool! I think that movie did really well when it was in the box office so it must be good!} \\
\textbf{Human 1}: & \textit{yea. what else do you like to do beside movies?} \\
\textbf{Human 2}: & \textit{I enjoy baking cookies. I am on a quest to bake the best chocolate chip cookie \Smiley What about you?} \\
\textbf{Human 1}: & \textit{I enjoy eating \Smiley} \\
\textbf{Human 2}: & \textit{so definitely would like to try your best chocolate cookie} \\
\textbf{Human 1}: & \textit{I will have to bake some soon and let you know. What types of food do you like to eat?} \\
\textbf{Human 2}: & \textit{thanks! I generally love noodle soups like Pho or Ramen :)} \\
\textbf{Human 1}: & \textit{Noodle soup is delicious! Do you make homemade noodle soup or do you prefer to go out?} \\
\textbf{Human 2}: & \textit{I prefer to go out. I'm not a good cook haha} \\
\textbf{Human 1}: & \textit{Same! Even though I bake, I cannot cook} \\
\textbf{Human 2}: & \textit{seems like we share a thing in common, yay!} \\
\bottomrule
\end{tabular}
\end{table}

\begin{table}[tbh!]
\small
\centering
\begin{tabular}{rp{6cm}}
\toprule
\multicolumn{2}{c}{{\bf Human Conversation 9}} \\
\midrule
\textbf{Human 1}: & \textit{Hi!} \\
\textbf{Human 2}: & \textit{Good afternoon!} \\
\textbf{Human 1}: & \textit{How has your week been?} \\
\textbf{Human 2}: & \textit{So far so good. It is holiday season. So just chilling} \\
\textbf{Human 1}: & \textit{I think I'm getting sick with a cold \Sadey So you should chill on my behalf too cause I'm out the game for all of December.} \\
\textbf{Human 2}: & \textit{lol Sorry to hear that. Are you planning anything fun for December?} \\
\textbf{Human 1}: & \textit{Nothing exciting. I'll be posted up at home for the most part. I did a lot of travelling this year so my budget would have stopped me even if I wasn't sick.} \\
\textbf{Human 2}: & \textit{\Laughey} \\
\textbf{Human 1}: & \textit{Do you have big plans?} \\
\textbf{Human 2}: & \textit{Yes! I am going to Hawaii! This will be my first time visiting Hawaii. Really excited about it.} \\
\textbf{Human 1}: & \textit{I love Hawaii. It's a good place to be. I like going there cause it's humid so I never have to put on lotion.} \\
\textbf{Human 2}: & \textit{lol this is the first time I heard from a boy who cares about humidity and lotion. I cannot agree more. } \\
\textbf{Human 1}: & \textit{Brooooo!!! It's so important. When I got to California beaches I have to carry 3 litres of lotion for the whole day.} \\
\textbf{Human 2}: & \textit{\Laughey} \\
\bottomrule
\end{tabular}
\end{table}

\begin{table}[tbh!]
\small
\centering
\begin{tabular}{rp{6cm}}
\toprule
\multicolumn{2}{c}{{\bf Human Conversation 10}} \\
\midrule
\textbf{Human 1}: & \textit{Hi!} \\
\textbf{Human 2}: & \textit{Oh hello. Long time no talk. How's the day going for yuo?} \\
\textbf{Human 1}: & \textit{Very well, thanks for asking. How has your day been?} \\
\textbf{Human 2}: & \textit{Getting better. I just recovered from a cold. I got wet in the rain last week. Are you planning anything for the holidays?} \\
\textbf{Human 1}: & \textit{Glad to hear you're better. Sorry to hear you were sick. I was sick a couple of weeks ago with a bad cough. There's definitely a bug going around. Admit I just want to stay healthy for the holidays and plan to relax.} \\
\textbf{Human 2}: & \textit{Oh same here. I think relaxing at home should be counted among the best ways to enjoy the holidays.} \\
\textbf{Human 1}: & \textit{Definitely! I know a lot of folks travel for the holidays, but I'm happy to stay home myself!} \\
\textbf{Human 2}: & \textit{I'm getting there. Every year until last year, I tried to go somewhere for the Christmas / New Year, and then I got bored traveling. lol not sure if that means I'm getting old?} \\
\textbf{Human 1}: & \textit{Me too. Now I have folks come visit me for the holidays! But that's also tiresome..} \\
\textbf{Human 2}: & \textit{Are you doing any home decorating then?} \\
\textbf{Human 1}: & \textit{Yes! We set up an eco-friendly (i.e. fake) Christmas tree and put up some colorful LED lights which is very festive.} \\
\textbf{Human 2}: & \textit{I think I'm copying you. Me and my wife plan to decorate and Christmas tree too. We bought most of the decorative stuffs from the stores, but haven't yet to buy the tree.} \\
\textbf{Human 1}: & \textit{Buying a tree is a neat experience. I was torn between buying an artificial/eco-friendly/fake one vs. a real one that smells like fresh pine. In the end, we opted for the one that we can disassemble every year.} \\
\textbf{Human 2}: & \textit{I see. Artificial anything is better, from tree to intelligence, huh?} \\
\textbf{Human 1}: & \textit{Oh, very clever pun! I like it! Depends. I remember having real Christmas trees from childhood, but these days with climate change, I think not chopping down a tree just to decorate it and then throw it out in a month is the more responsible thing to do.} \\
\textbf{Human 2}: & \textit{I see. It's probably also cheaper. I'll buy an artificial one too. Do you have any suggestions for the store?} \\
\textbf{Human 1}: & \textit{Admit my favorite store is Target, plus they often have good deals.} \\
\textbf{Human 2}: & \textit{Ah that's great. My wife also likes Target a lot. She even made a Target credit card because she comes to that store very often. Okay thanks for the suggestion. I'll check out Target.} \\
\textbf{Human 1}: & \textit{Great, I hope you find a nice tree.} \\
\bottomrule
\end{tabular}
\end{table}

\FloatBarrier

\onecolumn

\section{Additional Sample Responses}
We show additional responses from the \meenafull{}, Cleverbot, and DialoGPT for 60 contexts that the authors wrote before generating responses. These contexts are also not in the MTB dataset.

\begin{longtable}{p{.3\textwidth}| p{.2\textwidth}| p{.2\textwidth} | p{.2\textwidth}}
   \toprule
 {\bf Sample Contexts} & {\bf Meena} & {\bf Cleverbot} & {\bf DialoGPT} \\
 \midrule
\endhead
\endfoot
\endlastfoot
 \aturn{What do you think of the Warriors this season?} \newline \bturn{It's too bad Steph broke his hand :(} \newline \aturn{Yeah... Now who do you think will reign supreme?} & \bturn{Probably the Spurs.} & \bturn{I don't know, I don't really care about things that don't affect me greatly.} & \bturn{I hope Curry, I think he's the most likable player in the NBA.} \\
\midrule
\aturn{What do you think of the Warriors this season?} \newline \bturn{It's too bad Steph broke his hand :(} & \aturn{It's been broken for a while now.} & \aturn{Can you prove you are not a bot?} & \aturn{I think it's the same one.} \\
\midrule
\aturn{What do you think of the Warriors this season?} & \bturn{I haven't watched a full game yet} & \bturn{Nope, I'm a guy.} & \bturn{I like the warriors! But they have the most talented bench in the league. We need to start some young guns to make our bench better.} \\
\midrule
\aturn{What are you up to?} \newline \bturn{Waiting for Christmas!} \newline \aturn{It's the best time of the year!} & \bturn{I know I can't wait!} & \bturn{I know!} & \bturn{It sure is. I love it.} \\
\midrule
\aturn{What are you up to?} \newline \bturn{Waiting for Christmas!} & \aturn{Oh, I know that feel! What did you get for Christmas?} & \aturn{But I know he'll appear.} & \aturn{What a great idea!!} \\
\midrule
\aturn{What are you up to?} & \bturn{I'm at work. Just woke up.} & \bturn{*faints*.} & \bturn{Nothing, you?} \\
\midrule
\aturn{Do you believe in AGI?} \newline \bturn{Hmm ... what is AGI?} \newline \aturn{Oh, it's Artificial General Intelligence} & \bturn{Yes. Of course. What else?} & \bturn{I do not believe you.} & \bturn{ohh. ok, thanks.} \\
\midrule
\aturn{Do you believe in AGI?} \newline \bturn{Hmm ... what is AGI?} & \aturn{Artificial General Intelligence} & \aturn{A meat pie.} & \aturn{Associative personality disorder. It's one of the causes of many things that can happen to you.} \\
\midrule
\aturn{Do you believe in AGI?} & \bturn{Why not? It exists, why not believe in it?} & \bturn{What do you mean by that?} & \bturn{Do you believe in god?} \\
\midrule
\aturn{What's your favorite sandwich and why?} \newline \bturn{I can't go past a banh mi, because it reminds me of my trip to Vietnam.} \newline \aturn{Oh, I'd love to go there one day!} & \bturn{It's an incredible place. I highly recommend visiting.} & \bturn{Do you like cheese?} & \bturn{It was really good.} \\
\midrule
\aturn{What's your favorite sandwich and why?} \newline \bturn{I can't go past a banh mi, because it reminds me of my trip to Vietnam.} & \aturn{Banh mi is the best Vietnamese sandwich.} & \aturn{Do you like the ocean?} & \aturn{It's like the food of the people, but more like the food of the people you love.} \\
\midrule
\aturn{What's your favorite sandwich and why?} & \bturn{Bacon egg and cheese on a bagel} & \bturn{Peanut butter and jelly*.} & \bturn{Pepperoni in a bread bowl, because I can't think of a better way.} \\
\midrule
\aturn{If you could live in any fictional world, which would you choose?} \newline \bturn{I would love to live in The Culture, from the Iain M Banks books.} \newline \aturn{Are those the ones that Elon Musk is really into?} & \bturn{No, those ones are SpaceX and Tesla Motors.} & \bturn{No we are 6 foot.} & \bturn{I don't know, do they not have enough space travel for him to travel there?} \\
\midrule
\aturn{If you could live in any fictional world, which would you choose?} \newline \bturn{I would love to live in The Culture, from the Iain M Banks books.} & \aturn{I was hoping someone would mention The Culture. If I had to pick one place to live, the Culture would be it without a doubt.} & \aturn{Me too.} & \aturn{Oh yes, I love them.} \\
\midrule
\aturn{If you could live in any fictional world, which would you choose?} & \bturn{Harry Potter, for sure.} & \bturn{Anime.} & \bturn{Earth. I want to make a good living.} \\
\midrule
\aturn{You always choose Seal's ``Kiss From A Rose" at karaoke, but honestly you don't have the voice for it.} \newline \bturn{Whoa! Way harsh, Tai!} \newline \aturn{Sorry, but I set a high standard at karaoke and I expect everyone to meet it.} & \bturn{Sorry, I'm not Seal or Tai.} & \bturn{Because you fall in love too often, that's what's the matter.} & \bturn{It's not a high standard, it's a high standard.} \\
\midrule
\aturn{You always choose Seal's ``Kiss From A Rose" at karaoke, but honestly you don't have the voice for it.} \newline \bturn{Whoa! Way harsh, Tai!} & \aturn{It's just an observation - it's not a criticism!} & \aturn{Tai?} & \aturn{Well, at least it's something I know for a fact.} \\
\midrule
\aturn{You always choose Seal's ``Kiss From A Rose" at karaoke, but honestly you don't have the voice for it.} & \bturn{That's not how you spell ``Every Breath You Take."} & \bturn{I love you to the end of forever and back.} & \bturn{Haha. I've been trying to work out how to sing that.} \\
\midrule
\aturn{How did you become such a snappy dresser?} \newline \bturn{I studied to be a fashion designer, did you not know that?} \newline \aturn{No, I didn't! Wow.} & \bturn{It's because I'm the best!} & \bturn{Lol well now you do.} & \bturn{It's ok.It's like the third time this has happened and the third time they've said I've been doing that.} \\
\midrule
\aturn{How did you become such a snappy dresser?} \newline \bturn{I studied to be a fashion designer, did you not know that?} & \aturn{Did you also study to be a doctor?} & \aturn{Well now you do.} & \aturn{Oh... I did not. Thanks.} \\
\midrule
\aturn{How did you become such a snappy dresser?} & \bturn{I think it's the suit.} & \bturn{I am not a dancer.} & \bturn{It's a lifestyle.} \\
\midrule
\aturn{Are you any good at ping pong?} \newline \bturn{I practise a lot, but I don't seem to be getting better.} \newline \aturn{Don't be modest, I'm sure you must be great!} & \bturn{Thanks! I'll keep practising.} & \bturn{Do you believe in God?} & \bturn{Oh, I am definitely not!} \\
\midrule
\aturn{Are you any good at ping pong?} \newline \bturn{I practise a lot, but I don't seem to be getting better.} & \aturn{I hear you can get a lot of practice in at the local Chinese buffet.} & \aturn{Me neither. Do you exercise?} & \aturn{You need to practice. You just need to practice more.} \\
\midrule
\aturn{Are you any good at ping pong?} & \bturn{No. No I am not.} & \bturn{Yes.} & \bturn{I can do the ping pong part pretty good, yes. I've done a lot of it in my life.} \\
\midrule
\aturn{Do you know DOTA?} \newline \bturn{Yea, it's what I do in college.} \newline \aturn{Same here! Do you ever imagine being beaten by an AI computer?} & \bturn{Oh, for sure. It's bound to happen eventually.} & \bturn{I suppose so. I am human so most of my thoughts are from the human perspective.} & \bturn{Not really, I'd be more intimidated by an AI robot than a person.} \\
\midrule
\aturn{Do you know DOTA?} \newline \bturn{Yea, it's what I do in college.} & \aturn{Favorite hero?} & \aturn{I got it now.} & \aturn{Oh okay then.} \\
\midrule
\aturn{Do you know DOTA?} & \bturn{I do know DOTA, I prefer HoN though.} & \bturn{Yeah I know dota.} & \bturn{No, I don't know much about that game.} \\
\midrule
\aturn{Love at first sight} \newline \bturn{And ...?} \newline \aturn{That's what happened in my life.} & \bturn{And what happened after?} & \bturn{No one told you when to run, you missed the starting gun.} & \bturn{Well, it does seem like a fun way to get to know other people, and maybe you could be a good friend.} \\
\midrule
\aturn{Love at first sight} \newline \bturn{And ...?} & \aturn{The rest is up to you} & \aturn{That's it...} & \aturn{It's a joke that you made, that's why} \\
\midrule
\aturn{Love at first sight} & \bturn{I can't say that I would've expected this.} & \bturn{I... Don't know.} & \bturn{Thanks! :D} \\
\midrule
\aturn{I saw an amazing dream last night!} \newline \bturn{Oh yeah! What did you see} \newline \aturn{I was driving a Ferrari!} & \bturn{Wow! What type of Ferrari was it?} & \bturn{Oh?} & \bturn{Nice. I saw one too} \\
\midrule
\aturn{I saw an amazing dream last night!} \newline \bturn{Oh yeah! What did you see?} & \aturn{The last episode of The Sopranos!} & \aturn{Talk to you.} & \aturn{A dream that I had about a dog and a girl} \\
\midrule
\aturn{I saw an amazing dream last night!} & \bturn{What was your dream?} & \bturn{I can pretend what it feels like to have a nose and smell with a nose!} & \bturn{I saw one last night!} \\
\midrule
\aturn{Mirror Mirror on the wall...} \newline \bturn{Who is the most amazing of them all?} \newline \aturn{You are :)} & \bturn{Nuh uh, you!} & \bturn{Flirt :).} & \bturn{I love you!!} \\
\midrule
\aturn{Mirror Mirror on the wall...} \newline \bturn{Who is the most amazing of them all?} & \aturn{You! :D} & \aturn{Yeah tunnel vision on the floors.} & \aturn{The one and only.} \\
\midrule
\aturn{Mirror Mirror on the wall...} & \bturn{who's the fairest in the land?} & \bturn{Tell who is the greatest of them all.} & \bturn{What about meeeeeee?} \\
\midrule
\aturn{Wait! Don't go there} \newline \bturn{Why?} \newline \aturn{It's dark and spooky!} & \bturn{What if I want to be spooked?} & \bturn{Do you like dark and scary stuff?} & \bturn{I know.} \\
\midrule
\aturn{Wait! Don't go there} \newline \bturn{Why?} & \aturn{I went there when I was in 3rd grade. Haven't been back since.} & \aturn{I want to talk to you more.} & \aturn{I think he was being ironic.} \\
\midrule
\aturn{Wait! Don't go there} & \bturn{I have no idea what's going on here, but I'm scared.} & \bturn{Why not?} & \bturn{I'm going there for work...} \\
\midrule
\aturn{What is your favorite color?} \newline \bturn{Blue!} \newline \aturn{I feel majority people choose it, don't you?} & \bturn{I think blue is more popular than people think.} & \bturn{Yes. the world is at mercy of humans.} & \bturn{I like blue too, it just looks weird to me.} \\
\midrule
\aturn{What is your favorite color?} \newline \bturn{Blue!} & \aturn{Are you a member of the Blue Man Group?} & \aturn{Mine is Purple. :).} & \aturn{That's pretty cool.} \\
\midrule
\aturn{What is your favorite color?} & \bturn{Blue. What's yours?} & \bturn{I don't.} & \bturn{I love it! It's a combination of green, blue and red.} \\
\midrule
\aturn{What came first: Egg or Chicken?} \newline \bturn{Chicken!} \newline \aturn{And what's your reasoning behind that?} & \bturn{The egg was laid by something that was not quite a chicken but was very close. So the egg must have come first!} & \bturn{There is no reasoning.} & \bturn{Because it's more efficient in the long run.} \\
\midrule
\aturn{What came first: Egg or Chicken?} \newline \bturn{Chicken!} & \aturn{The chicken or the egg that hatches the chicken?} & \aturn{How?} & \aturn{Chicken, or chicken?} \\
\midrule
\aturn{What came first: Egg or Chicken?} & \bturn{Egg, dinosaurs laid eggs.} & \bturn{The egg.} & \bturn{The egg came first, it's a chicken. The egg came after the chicken came.} \\
\midrule
\aturn{Do you like spicy food?} \newline \bturn{Yeah I love spicy food, specially Chinese} \newline \aturn{From what province?} & \bturn{From Guangdong province} & \bturn{Secret.} & \bturn{Hong Kong and I love chinese food} \\
\midrule
\aturn{Do you like spicy food?} \newline \bturn{Yeah I love spicy food, specially Chinese} & \aturn{How do you feel about Korean food?} & \aturn{Where are you from?} & \aturn{Chinese food is the bomb} \\
\midrule
\aturn{Do you like spicy food?} & \bturn{I love spicy food.} & \bturn{I love spicy food.} & \bturn{I think it was meant as a rhetorical question, which I found very funny.} \\
\midrule
\aturn{Have you ever tried a mechanical keyboard?} \newline \bturn{Yeah. My keyboard is mechanical. I really like how it sounds when I type} \newline \aturn{Yeah me too} & \bturn{What kind(s) do you have?} & \bturn{What is your favorite animal?} & \bturn{Me 3} \\
\midrule
\aturn{Have you ever tried a mechanical keyboard?} \newline \bturn{Yeah. My keyboard is mechanical. I really like how it sounds when I type} & \aturn{What switches?} & \aturn{Gee it's swell to finally beat her other...} & \aturn{Have you tried rubber dome keyboards?} \\
\midrule
\aturn{Have you ever tried a mechanical keyboard?} & \bturn{Yes. Once you go clack, you never go back.} & \bturn{Yes stupid.} & \bturn{No. Not yet. But I'm going to give it a try.} \\
\midrule
\aturn{Do you have any new year resolution?} \newline \bturn{I want to get a six pack} \newline \aturn{Cool! What's your motivation?} & \bturn{I want to be more fit} & \bturn{I have already told you that.} & \bturn{To be more active in my workouts} \\
\midrule
\aturn{Do you have any new year resolution?} \newline \bturn{I want to get a six pack} & \aturn{What are you doing to work on that?} & \aturn{What kind of job?} & \aturn{I like you. This is what's wrong with the world.} \\
\midrule
\aturn{Do you have any new year resolution?} & \bturn{Get good grades.} & \bturn{No.} & \bturn{No. I just want to get the money out and get a better paying job so that I can buy a new computer next year.} \\
\midrule
\aturn{I think I'm going to eat cake} \newline \bturn{Can I come with you?} \newline \aturn{Sure! Do you like French cakes?} & \bturn{I love French cakes.} & \bturn{Of course I do. I just said that.} & \bturn{They're great.} \\
\midrule
\aturn{I think I'm going to eat cake} \newline \bturn{Can I come with you?} & \aturn{Of course! Cake buddies!} & \aturn{Of course you can.} & \aturn{Come for cake. Stay for the rest of the game.} \\
\midrule
\aturn{I think I'm going to eat cake} & \bturn{What kind of cake? :)} & \bturn{WHat do you want to eat?} & \bturn{What kind?} \\
\midrule
\aturn{Do you speak any other languages?} \newline \bturn{I speak French} \newline \aturn{Nice. Have you been to France?} & \bturn{Non, tes-vous fran ais?} & \bturn{No.} & \bturn{Yeah, but only for 2 days} \\
\midrule
\aturn{Do you speak any other languages?} \newline \bturn{I speak French} & \aturn{Parlez vous fran ais?} & \aturn{Tell me something in french.} & \aturn{I have a friend in Paris.} \\
\midrule
\aturn{Do you speak any other languages?} & \bturn{I speak English.} & \bturn{I speak some Spanish.} & \bturn{Not at all, I just really want to learn some more languages} \\
\bottomrule 
\caption{Additional sample responses for Meena, Cleverbot, and DialoGPT.\label{t:additional_samples}}
\end{longtable}

\begin{table*}[tbh!]
\small
  \centering
  \resizebox{16cm}{!}{
  \begin{tabular}{p{6cm}|p{4cm}}
  \toprule
{\bf Sample Contexts} & {\bf GenericBot} \\
\midrule
% 3-turn contexts
\aturn{You must have read a lot of books} \newline \bturn{Oh wow, how do you know? Do I look nerdy?} & \bturn{ok}  \\
\aturn{Of course, you're not!} & \rates{sensible: 1, specific: 0} \\
\cdashlinelr{1-2}
\aturn{You know nothing, Jon Snow!} \newline \bturn{You must be a fan of Game of Thrones}  & \bturn{I don't know} \\
\aturn{Haha yea, how can I not?} & \rates{sensible: 0, specific: 0} \\
\cdashlinelr{1-2}
\aturn{What is your favorite island in the world?} \newline \bturn{I would say Fiji} & \bturn{I don't know} \\
\aturn{Interesting, what's nice about it?} & \rates{sensible: 0, specific: 0} \\
\midrule
% 2-turn contexts
\multirow{2}{6cm}{\aturn{You must have read a lot of books} \newline \bturn{Oh wow, how do you know? Do I look nerdy?}} & \bturn{I don't know\\} \\
& \rates{sensible: 1, specific: 0} \\
\cdashlinelr{1-2}
\multirow{2}{6cm}{\aturn{You know nothing, Jon Snow!} \newline \bturn{You must be a fan of Game of Thrones}}  & \aturn{ok\\} \\
& \rates{sensible: 0, specific: 0} \\
\cdashlinelr{1-2}
\multirow{2}{6cm}{\aturn{What is your favorite island in the world?} \newline \bturn{I would say Fiji}} & \aturn{ok\\} \\
& \rates{sensible: 1, specific: 0} \\
\midrule
% 1-turn
\multirow{2}{6cm}{\aturn{You must have read a lot of books}} & \bturn{ok\\} \\
& \rates{sensible: 0, specific: 0} \\
\cdashlinelr{1-2}
\multirow{2}{6cm}{\aturn{You know nothing, Jon Snow!}} & \bturn{ok\\} \\
& \rates{sensible: 0, specific: 0} \\
\cdashlinelr{1-2}
\multirow{2}{6cm}{\aturn{What is your favorite island in the world?}} & \bturn{I don't know\\} \\
& \rates{sensible: 1, specific: 0} \\
\bottomrule
  \end{tabular}
  }
  \caption{{\bf Sample GenericBot responses from static evaluation} -- Shown are responses of GenericBot on some sample contexts from Table \ref{t:static_analysis} together with their binary labels per category (sensibleness and specificity). GenericBot responds to questions with ``I don't know" and to statements with ``ok". Note that the contexts were constructed in advance before we generate all bot responses. These contexts are not in the MTB dataset.}
  \label{t:boring_bot_static_analysis}
\end{table*}

\section{Additional Figures}

\begin{figure}[ht!]
\centering
\includegraphics[width=230pt]{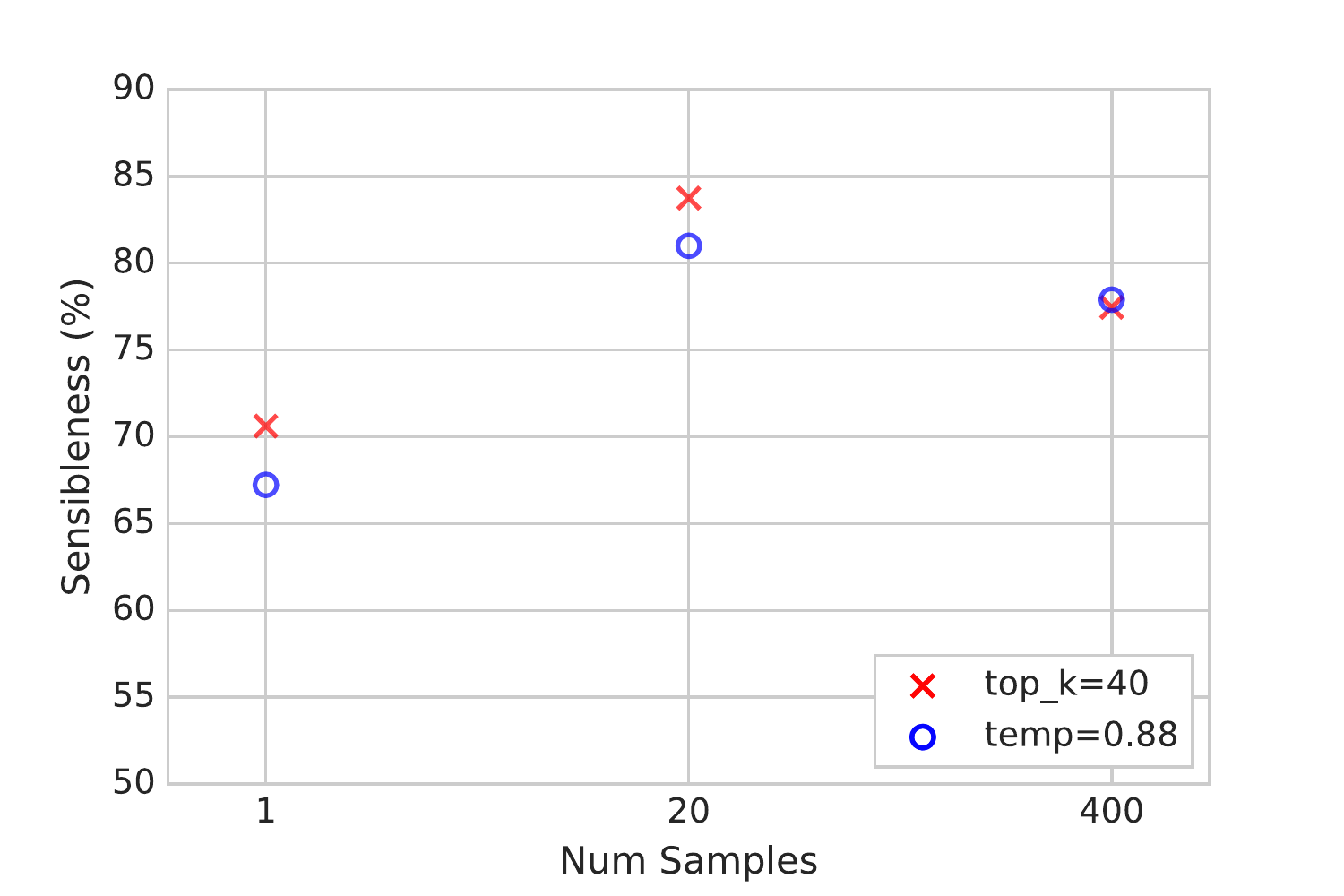}
\caption{Static sensibleness over number of sampled responses for top-\textit{k} and sampling with temperature.}
\label{fig:decoding-sensibleness}
\end{figure}

\begin{figure}[ht!]
\centering
\includegraphics[width=230pt]{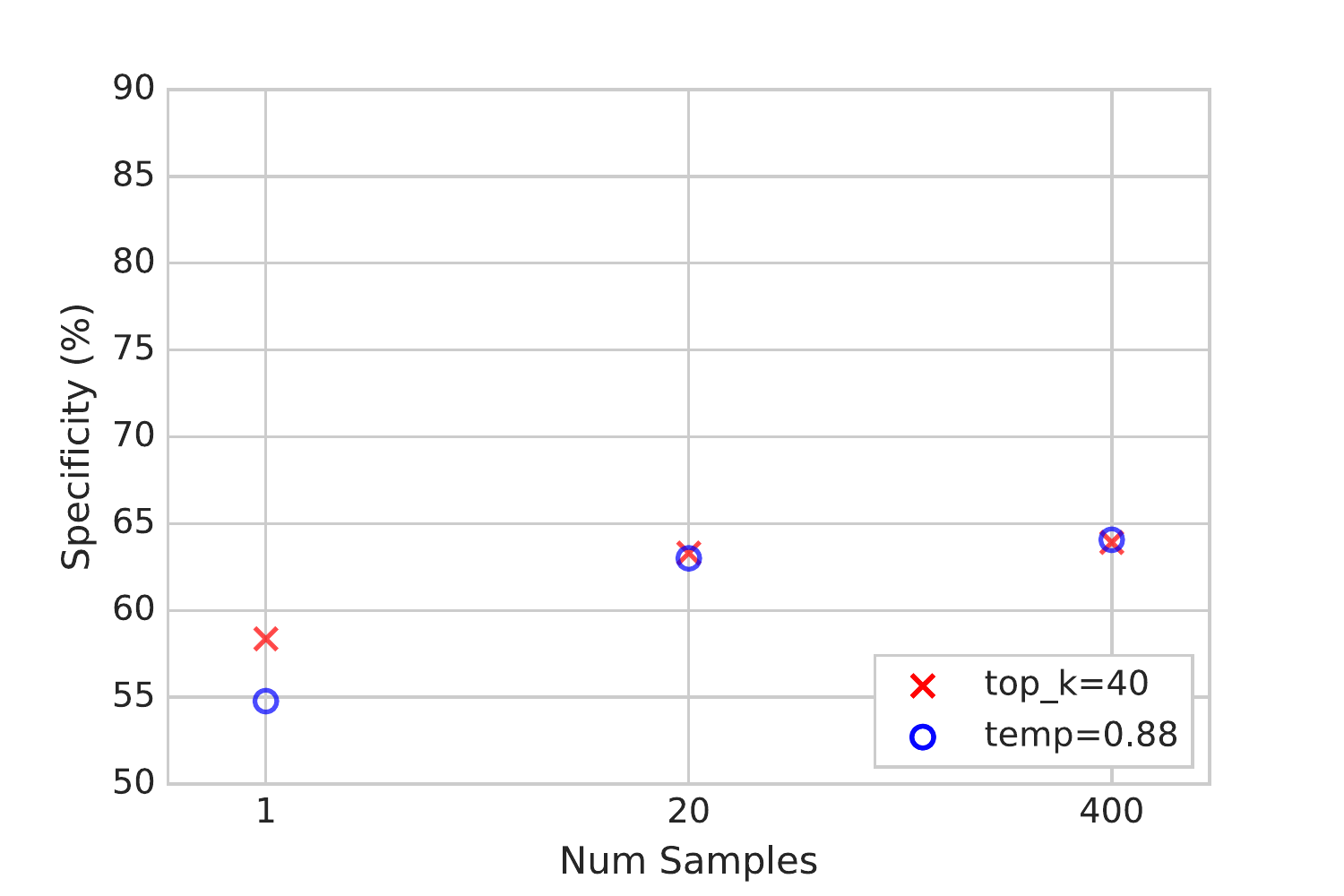}
\caption{Static specificity over number of sampled responses for top-\textit{k} and sampling with temperature.}
\label{fig:decoding-specificity}
\end{figure}
 
\begin{figure}[ht!]
\centering
\includegraphics[height=150pt]{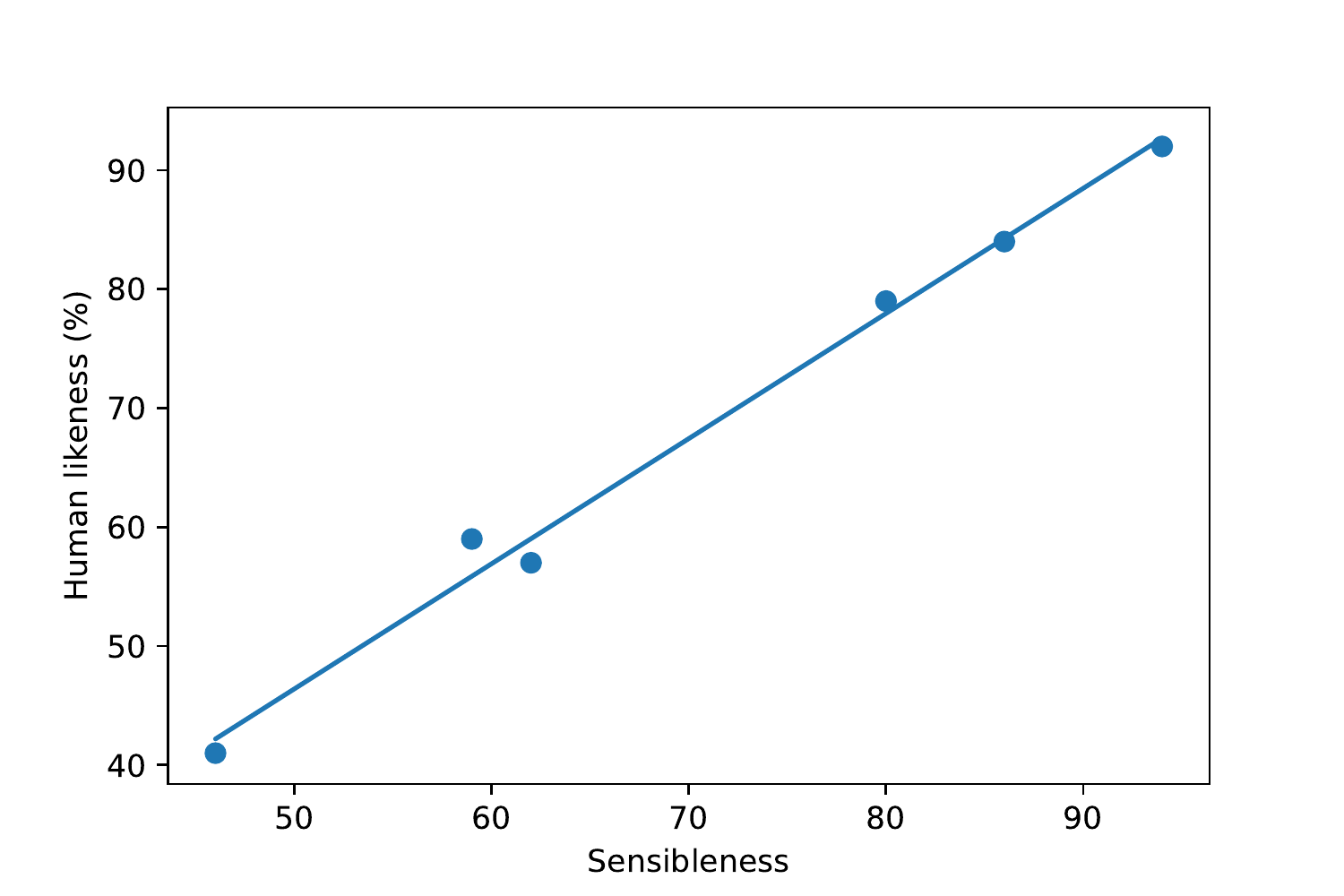}
\caption{Sensibleness vs human likeness. Each point is a different chatbot, except for the top right one, which is human. A regression line is plotted, for which the coefficient of determination ($R^2$) is 0.99, an indication of strong correlation between sensibleness and human likeness.}
\label{fig:sensibleness-human-likeness}
\end{figure}

\begin{figure}[ht!]
\centering
\includegraphics[height=150pt]{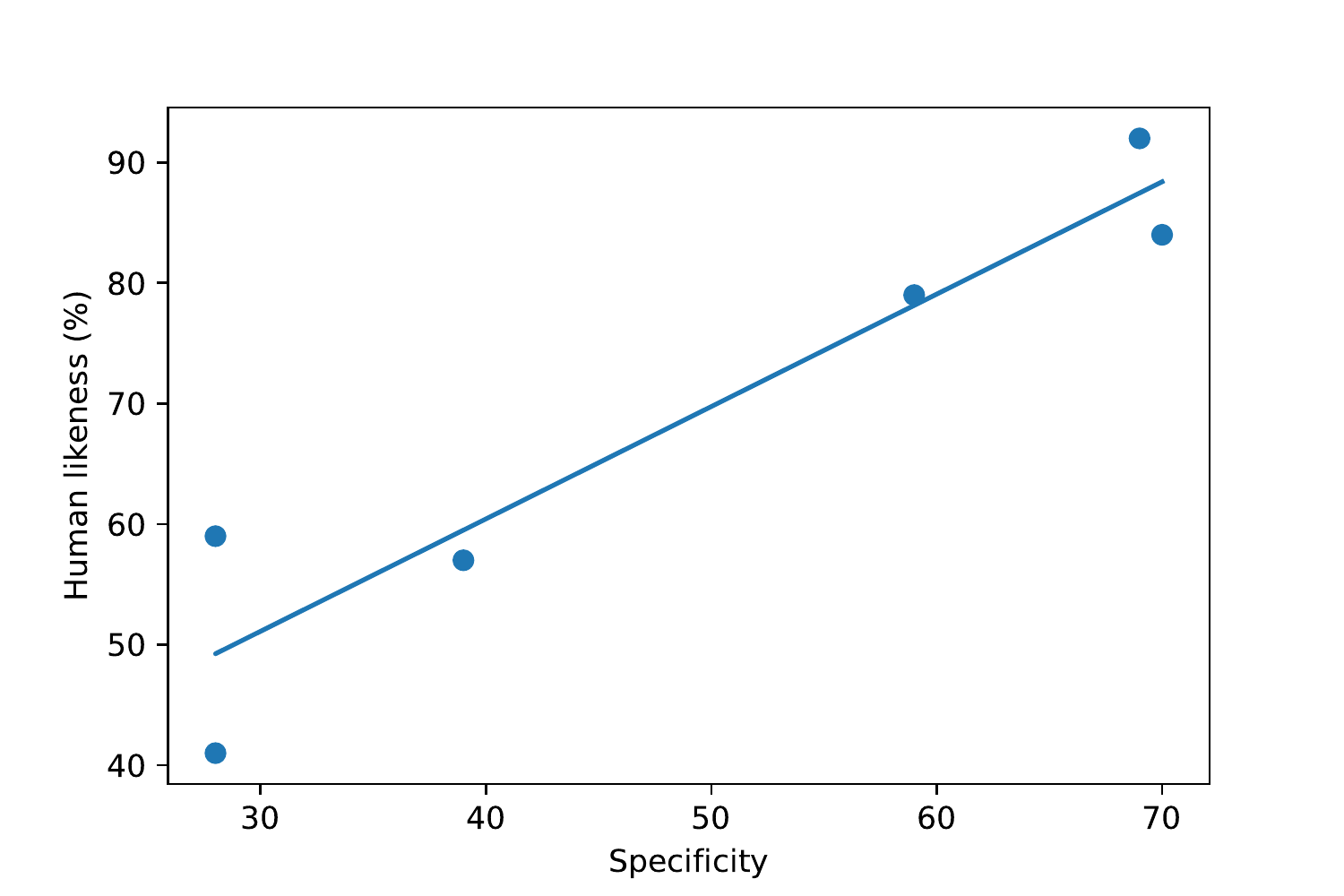}
\caption{Specificity vs human likeness. Each point is a different chatbot, except for the top right one, which is human. A regression line is plotted, for which the coefficient of determination ($R^2$) is 0.89, an indication of strong correlation between specificity and human likeness.}
\label{fig:specificity-human-likeness}
\end{figure}

\end{document}